\documentclass[10pt,twocolumn,letterpaper]{article}
\usepackage{cvpr}              % To produce the CAMERA-READY version
\usepackage[accsupp]{axessibility}
\usepackage{CJKutf8}
\usepackage{graphicx}
\usepackage{float}
\usepackage{array}
\usepackage{amsmath,amssymb,amsfonts}
\usepackage{booktabs}
\usepackage{tabularx}
\usepackage{multirow}
\usepackage{color}
\usepackage[pagebackref,breaklinks,colorlinks]{hyperref}
\usepackage{paralist}

\newcommand{\eref}[1]{Eq.~(\ref{#1})}
\newcommand{\fref}[1]{Figure~\ref{#1}}
\newcommand{\sref}[1]{Section~\ref{#1}}
\newcommand{\tref}[1]{Table~\ref{#1}}

\def\cvprPaperID{79} % *** Enter the CVPR Paper ID here
\def\confName{CVPR}
\def\confYear{2023}

\begin{document}

\title{StructVPR: Distill Structural Knowledge with Weighting Samples \\ for Visual Place Recognition}

\author{Yanqing Shen ~Sanping Zhou ~Jingwen Fu ~Ruotong Wang ~Shitao Chen ~Nanning Zheng\thanks{Corresponding author.}
\thanks{Supported by National Science Foundation of China (No. 62088102).}\\
National Key Laboratory of Human-Machine Hybrid
Augmented Intelligence, \\ 
National Engineering Research Center for Visual Information and Applications, \\ 
and Institute of Artificial Intelligence and Robotics, Xi'an Jiaotong University\\
% Institute of Artificial Intelligence and Robotics, Xi’an Jiaotong University\\
% {\small \tt \{qing1159364090@stu., spzhou@, fu1371252069@stu., wrt072@stu., chenshitao@, nnzheng@mail.\}xjtu.edu.cn}
}

\begin{CJK*}{UTF8}{gbsn}
\CJKindent
% 进度页
% red/blue/cyan
% \underline

\maketitle

\begin{abstract}
Visual place recognition (VPR) is usually considered as a specific image retrieval problem.
Limited by existing training frameworks, most deep learning-based works cannot extract sufficiently stable global features from RGB images and rely on a time-consuming re-ranking step to exploit spatial structural information for better performance.
In this paper, we propose StructVPR, a novel training architecture for VPR, to enhance structural knowledge in RGB global features and thus improve feature stability in a constantly changing environment.
Specifically, StructVPR uses segmentation images as a more definitive source of structural knowledge input into a CNN network and applies knowledge distillation to avoid online segmentation and inference of seg-branch in testing.
Considering that not all samples contain high-quality and helpful knowledge, and some even hurt the performance of distillation,
we partition samples and weigh each sample's distillation loss to enhance the expected knowledge precisely.
Finally, StructVPR achieves impressive performance on several benchmarks using only global retrieval and even outperforms many two-stage approaches by a large margin.
After adding additional re-ranking, ours achieves state-of-the-art performance while maintaining a low computational cost.
\end{abstract}

\section{Introduction}
% 重中之重,逻辑清楚
% 需要讲清楚为什么是structural knowledge【GV】

% 【1】\textbf{X is important}
Visual place recognition (VPR) is a critical task in autonomous driving and robotics, and researchers usually regard it as an image retrieval problem\cite{lowry2015visual,zhang2021visual,masone2021survey}. 
Given a query RGB image from a robot, VPR aims to determine whether the robot has been to this place before and to identify the corresponding images from a database.
Extreme environmental variations are challenging to methods, especially long-term changes (seasons, illumination, vegetation) and dynamic occlusions.
Therefore, learning discriminative and robust features is essential to distinguish places.

\begin{figure}[t]
  \centering
\subfloat[SEG is better than RGB]{
  \label{fig:seg}
  \includegraphics[width=0.47\columnwidth]{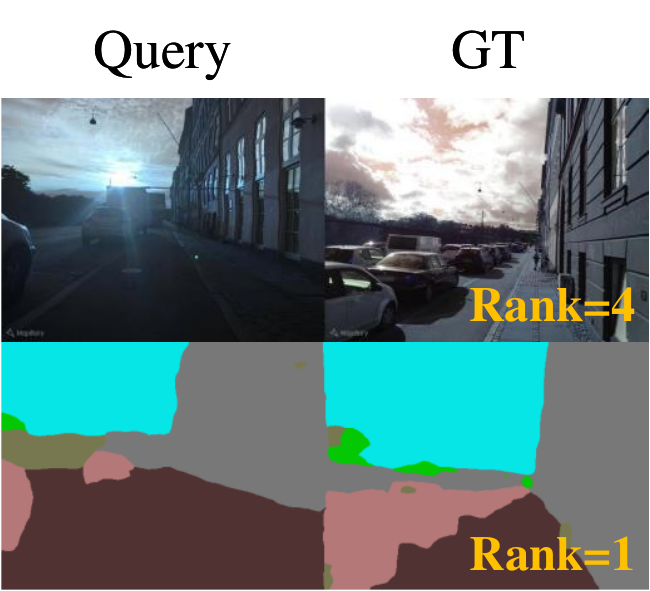}}
\subfloat[RGB is better than SEG]{
  \label{fig:rgb}
  \includegraphics[width=0.47\columnwidth]{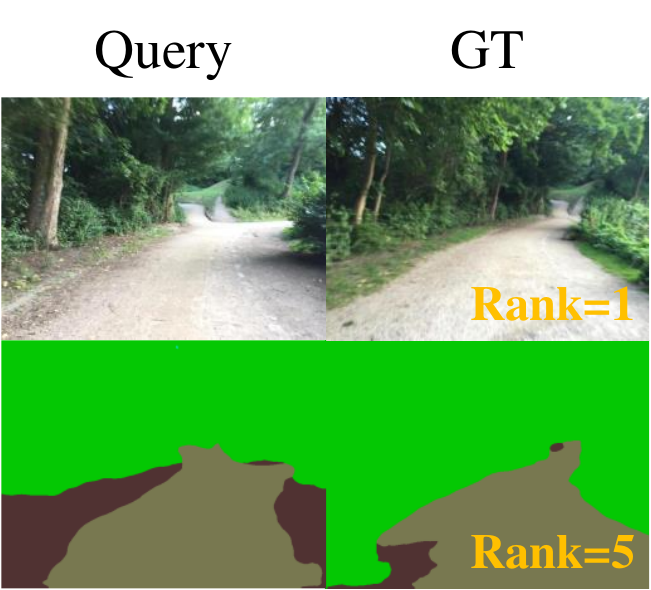}}
  \vspace{-0.2cm}
   \caption{\textbf{Examples of query images and ground truths.} The marked number represents the recall performance of two pre-trained branches on ground truths. (a) shows the scene with illumination variation and seasonal changes, where segmentation images are more recognizable. (b) shows the scene with changing perspectives, where RGB images are more recognizable.}
   \label{fig:show}
\end{figure}

% 【2】\textbf{A,B,C have been done; have their weakness; analyze the weakness}% 需要介绍经典VPR、引入segmentation、depth的VPR、经典的KD处理
%notes 不是不丰富 而是不显式
% notes 强调long-term
{There has been a commonly used two-stage strategy that retrieves candidates with global features and then re-ranks them through local descriptor matching, where re-ranking is time- and resource-consuming but dramatically improves the recall performance.
The improvement is because geometric verification based on local features provides rich and explicit structural information, which has stronger robustness to VPR than appearance information in some aspects, such as shape, edge\cite{wasabi}, spatial layout, and category\cite{xview}.}
% 之前也有人用segmentation:比如多任务、幻觉网络等---更偏semantic
Considering that segmentation (SEG) images have rich structural information, we tried some empirical studies using RGB and SEG as network input for VPR.
{We find that these two modalities have their advantages and disadvantages at the sample level,}
which means that better performance can be achieved if both modalities are appropriately fused. As \fref{fig:show} shows, both modalities have specific cases they are better at recognizing.

%这是引入segmentation的必要性，也就是的确有优势
% 互补和蒸馏之间的关系 这意味着使用得当就可以实现更好的精度互补 不是绝对的弱于RGB

% 【3】\textbf{how to solve; our work D}
% 用两者信息--确定蒸馏的这种框架--确认蒸馏细节
% 我们的目标是enhance rich structural information in feature representation, 用全局替代rerank
% 定下前提：只用全局和RGB，实现SOTA
{Based on the above discussion, we attempt to use the SEG modality to enhance structural knowledge in global RGB feature representation, achieving comparable performance to re-ranking while maintaining low computational cost.}
Complementarity of the two modalities on samples, that is, some samples may contain harmful knowledge for RGB, inspires us to perform knowledge enhancement selectively.
% Meanwhile, to avoid online generation and inference of segmentation images during testing, offline knowledge distillation (KD) is used here.
% multi-task framework or knowledge distillation can be used. But compared with multi-task framework, knowledge distillation is more direct and interpretable for enhancing structural information in features, so offline knowledge distillation (KD) is used here.
% 针对segmentation的处理% 需要借助蒸馏% 样本划分嵌入到蒸馏里
% Furthermore, the teacher network (SEG) is not absolutely stronger than the student network (RGB), which makes some samples hurt the whole effect of knowledge distillation.
Therefore, we propose a new knowledge distillation (KD) architecture, StructVPR, which can effectively distill the high-quality structural representations from the SEG modality to the RGB modality.
Specifically, StructVPR uses RGB images and encoded segmentation label maps for separate pre-training with VPR loss, uses two pre-trained branches to partition samples, and then weights the distillation loss to selectively distill the high-quality knowledge of the pre-trained seg-branch into the final RGB network.
Note that the concept of ``sample'' is a sample pair with a query and a labeled positive.
Compared with non-selective distillation methods\cite{dai2021learning} and previous selective distillation works\cite{selective}, StructVPR can exactly mine those suitable samples on which the teacher network performs good and better than the student and distinguish the importance of sample knowledge for KD.
Moreover, overly refined segmentation is not helpful, and the importance of all semantic classes varies for VPR.
Hence, we cluster original classes according to the sensitivity to objects in VPR and introduce prior information of labels via weighted one-hot encoding.

% beneficial
%   while preserving segmentation information learned structural representations and used as network input 
Our main contributions can be highlighted as follows:
    1) The overall architecture avoids the computation and inference of segmentation during testing by distilling the high-quality knowledge from the SEG modality to the RGB modality, where segmentation images are pre-encoded into weighted one-hot label maps to extract structural information for VPR.
    2) To the best of our knowledge, there is no previous work in VPR concerning selecting suitable samples for distillation. StructVPR forges a connection between sample partition with student network participating and weighted knowledge distillation for each sample.
    % 连续的权重分配
    % the selection of suitable sample pairs for weighted distillation.
    % sample partition and selectively learn the knowledge across suitable sample pairs.  to build descriptors.
    % 3) In pre-training stage, segmentation images are encoded into weighted one-hot label maps via class clustering and used as network input, which is more conducive to extracting effective structural information for VPR.
    % \item We demonstrate the sample-level complementarity of RGB images and segmentation images for VPR.
    3) We perform comprehensive experiments on key benchmarks. StructVPR performs better than global methods and achieves comparable performance to many two-stage (global-local) methods. The consistent improvement in all datasets corroborates the effectiveness and robustness of StructVPR.
Experimental results show that StructVPR achieves SOTA performance with low computational cost compared with global methods, and it is also competitive with most two-stage approaches \cite{delg,hausler2021patch}.
StructVPR with re-ranking outperforms the SOTA VPR approaches (4.475$\%$ absolute increase on Recall$@5$ compared with the best baseline\cite{transvpr}).

\begin{figure*}
    \centering
    \includegraphics[width=\linewidth]{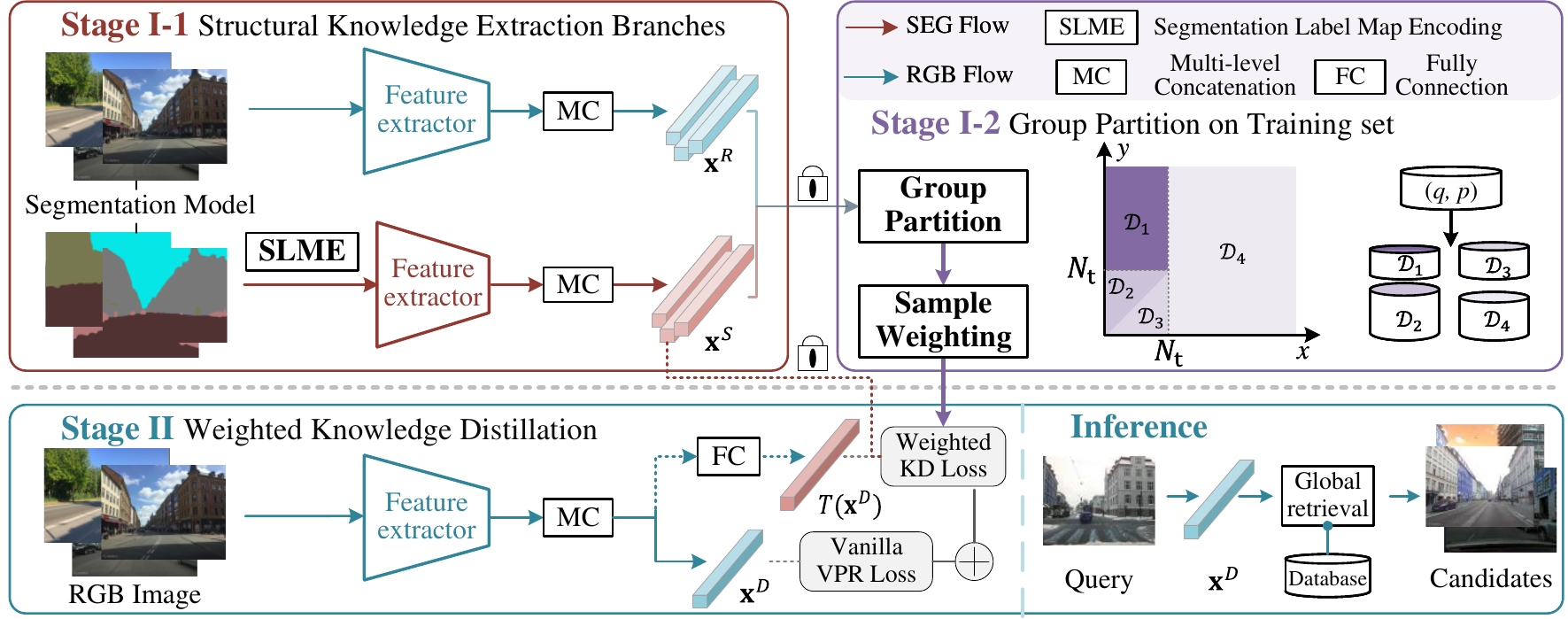}
   \vspace{-0.5cm}
    \caption{\textbf{Illustration of the proposed pipeline.} We first pre-train two branches with VPR supervision to extract structural knowledge, and then we perform offline group partition and weighting on samples. Then weighted knowledge distillation and VPR supervision are performed in Stage II. During testing, StructVPR avoids the computation and inference of segmentation, using only the trained model in Stage II. More importantly, it can maximize the efficiency of distilling high-quality knowledge.}
    \label{fig:pipeline}
\end{figure*}
% \textcolor[rgb]{0.1,0.6,0.7}{Blue lines} refer to the RGB flow and \textcolor[RGB]{202,12,22}{red lines} refer to the segmentation (SEG) flow. 
% the partitioning and weighting mechanisms from the sample perspective

\section{Related Work}

% 避免引用太长；去掉重复的
\textbf{Visual Place Recognition.} 
Most of the research on VPR has focused on constructing better image representations to perform retrieval. 
The common way to represent a single RGB image is to use global descriptors or local descriptors.
More recently, using CNNs to build local descriptors has achieved superior performances\cite{noh2017large,dusmanu2019d2,khaliq2019holistic,camara2020visual}.
Global descriptors can be generated by directly extracting\cite{zhou1,chen2017deep,gordo2017end,radenovic2018fine,revaud2019learning} or aggregating local descriptors. 
For aggregation, traditional methods have been incorporated into CNN-based architectures \cite{netvlad, mohedano2016bags}.
To achieve a good compromise between accuracy and efficiency, a widely used architecture is to rank the database by global features, and then re-rank the top candidates\cite{sta,sarlin2020superglue}. Many studies have verified its validity \cite{sarlin2019coarse,schuster2019sdc,teichmann2019detect,tcl}. 

Recently, many methods \cite{piasco2019learning,piasco2021improving,peng2021semantic,zhou2,zhou3} try to introduce semantics into RGB features by using attention mechanism or additional information. % 归属
TransVPR\cite{transvpr} introduces the attention mechanism to guide models to focus on invariant regions and extract robust representations.
{DASGIL\cite{dasgil} uses multi-task architecture with a single shared encoder to create global representation, and uses domain adaptation to align models on synthetic and real-world datasets.
Based on \cite{dasgil}, \cite{paolicelli2022learning} focuses on filtering semantic information via an attention mechanism.}
% label semtnaic; structural knowledge；pixel-wise
% todo 是不是表达出了这些意思；多任务的本质还是通过共享的encoder，从重建semantic segmentation的过程中回传对VPR有用的信息；segvpr则是利用了一种类似mask的机制去筛选；我们则是直接使用网络提取seg做VPR任务的特征信息

% Differently, our work uses a seg-branch to directly extract structural knowledge from segmentation images, and enhance them in RGB representations through weighted KD.

\textbf{Knowledge Distillation.} 
% 经典的模型压缩
% 多模态的KD
% 选择性KD
It is an effective way to enrich models with knowledge distillation (KD)\cite{gou2021knowledge}. 
It extracts specific knowledge from a stronger model (i.e., ``teacher'') and transfer to a weaker model (i.e., ``student'') through additional training signals.
There has been a large body of work on transferring knowledge with the same modality, such as model compression\cite{bucilu2006model,chen2017learning,hinton2015distilling} and domain adaptation\cite{li2017large,asami2017domain}. 
However, the data or labels for some modalities might not be available during training or testing, so it is essential to distill knowledge between different modalities \cite{ren2021learning,zhao2020knowledge}.
\cite{garcia2018modality,hoffman2016learning} generate a hallucination network to model depth information and enforce it for RGB descriptors learning. In this way, the student learns to simulate a virtual depth that improves the inference performance.
In this work, we construct a weighted knowledge distillation architecture to distill and enhance high-quality structural knowledge into RGB features.

Nevertheless, these previous works enhance the target model by transferring knowledge on each training sample from the teacher model, rarely discussing the difference about knowledge among samples \cite{liang2021reinforced,ge2018low}.
% 扩充选择蒸馏
Wang \textit{et al.} \cite{selective} proposes to select suitable samples for distillation through analyzing the teacher network.
Differently, our solution considers both pre-trained teacher and student network in sample partition and weight the distillation loss for samples.
% 第一个用于VPR的蒸馏 the domain knowledge in VPR

% \textcolor{red}{In addition, considering that VPR is essentially to improve the model robustness to the changes contained in sample pairs, the ``samples'' we partition are the sets of pairs consisting of a query and a positive.}

%notes 我们关心的是同一场景、不同拍摄之间的差异对应的模态知识，哪个模态更擅长

\section{Methodology}

{Considering that re-ranking with local features can provide rich and explicit structural knowledge, we try to utilize segmentation images to embed rich structural knowledge into RGB global retrieval and achieve comparable performance to two-stage methods.}
The key idea of StructVPR is very general: selectively distill high-quality and helpful knowledge into a single-input model by weighting samples. StructVPR includes two training stages: structural knowledge extraction and group partition, and weighted knowledge distillation. \fref{fig:pipeline} explains how StructVPR works. 

\subsection{Overview}
\label{subsec:over}
Given an input RGB image $I^{R}$, its pairwise segmentation image $I^{S}$ is extracted by an open-source semantic segmentation model and converted into a fixed-size label map via \textit{segmentation label map encoding} (SLME) module.

Then the seg-branch and rgb-branch are pre-trained with VPR loss.
Based on the two pre-trained branches and designed evaluation rules, training sets can be partitioned, and different groups represent samples with different performances of seg-branch and rgb-branch. 
Given the domain knowledge in VPR, the concept of ``sample'' is generalized as \textit{a sample pair of a query and a positive}.
There are two specific aspects of domain knowledge:
First, VPR datasets are often organized into \textit{query} and \textit{database}, in which each query image $q$ has a set of positive samples $\{p^{q}\}$ and a set of negative samples $\{n^{q}\}$.
Second, the knowledge to be enhanced is feature invariance and robustness under changes between sample pairs in a scene.

Moreover, a weighting function is defined, and we perform weighted knowledge distillation and vanilla VPR supervision in Stage II.

\subsection{Segmentation Label Map Encoding}
\label{subsec:encoding_module}
% 编码步骤示意？主要完成了3件事情
% 经过三个步骤，我们可以实现对分割信息到标准输入的编码
% 达成目标：降低计算、加速收敛、提升信息提取能力、引导、更干净的标签
It is nontrivial to convert segmentation images to a standard format before inputting them into the seg-branch, as each image contains a different number of semantic instances.
As shown in \fref{fig:pipeline}, the segmentation label map encoding (SLME) function encodes segmentation information to standard inputs of CNNs, including three steps: formatting, clustering, and weighting.

% 标准大小的格式 填充0-1
Firstly, like LabelEnc\cite{hao2020labelenc}, we use a $C \times H \times W$ tensor to represent segmentation images, where $H \times W$ equals the RGB image size and $C$ is the number of semantic classes.
Regions of the c-th class are filled with positive values in the c-th channel and 0 in other channels.

\begin{figure}
    \centering
    \includegraphics[width=0.9\linewidth]{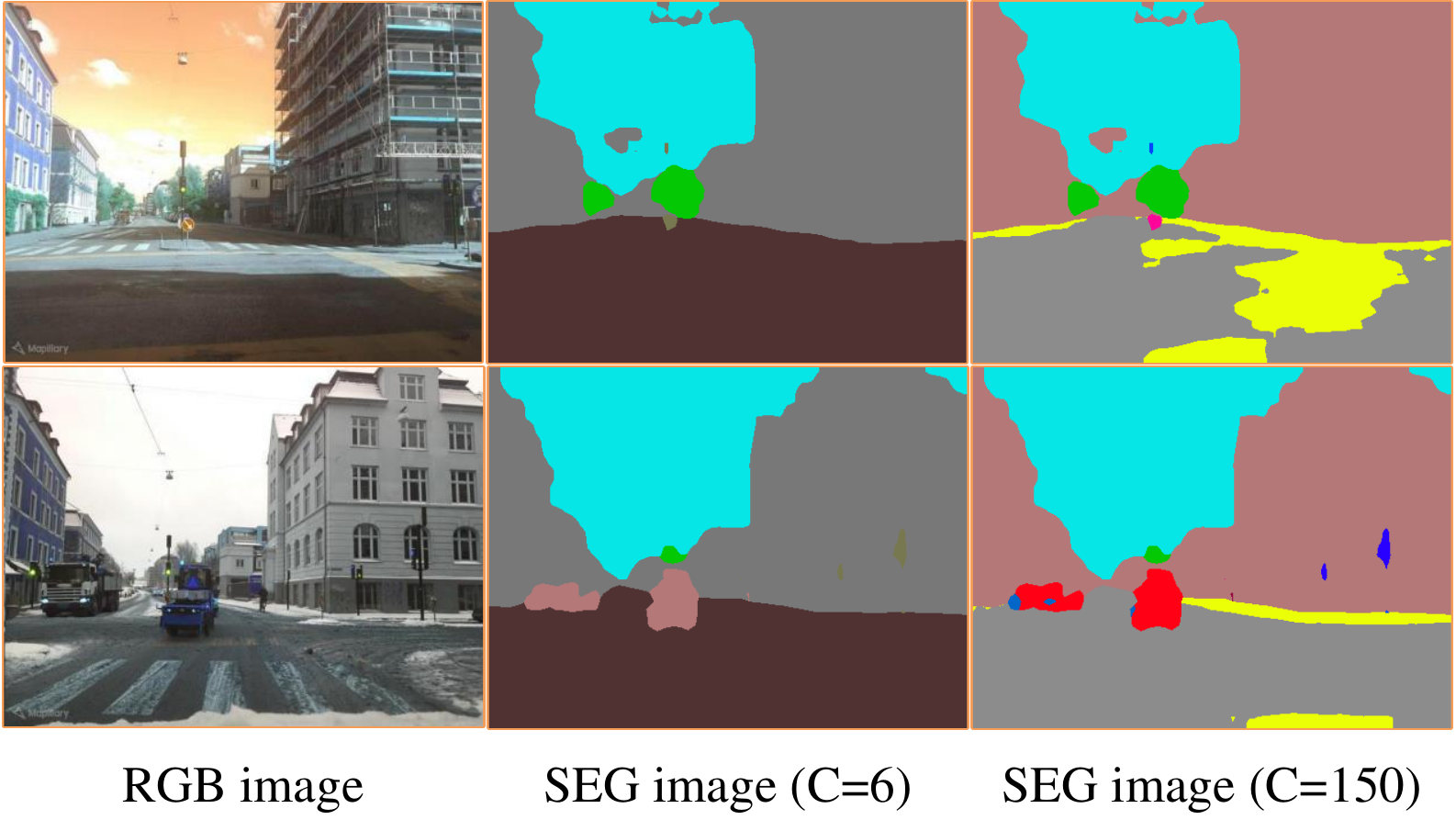}
   \vspace{-0.2cm}
    \caption{\textbf{Visualization of clustered semantic classes.} Shown from left to right are RGB images, 6-class segmentation images, and original segmentation images. It can be seen that the label space after clustering is cleaner for VPR.}
    \label{fig:cluster}
\end{figure}

% 聚类
Secondly, derived from human experience, we incorporate the semantic labels that have similar effects on VPR, such as cars and bicycles being re-labeled as ``dynamic objects".
It does this because too fine-grained segmentation will interfere with VPR like noises\cite{paolicelli2022learning}, which distracts model's ``attention'' and increase the difficulty of model convergence, shown in \fref{fig:cluster}. Moreover, a large $C$ will lead to excessive computation. 
Moreover, this step lets models focus on useful parts and reduces computational costs.

% weighted 
Finally, based on the idea that each semantic class plays a different role in VPR task, we weigh positive values in the encoding as prior information to guide the optimization of models and accelerate their convergence.
The attention mechanism of the human brain will allow humans to focus on iconic objects, such as buildings, while ignoring dynamic objects, such as pedestrians. In~\sref{sec:ablation}, experimental results corroborate our idea.

\subsection{Structural Knowledge Extraction}
\label{subsec:branch}
In the first training stage, we train seg-branch and rgb-branch to extract structural knowledge into features and prepare for subsequent group partition.

\textbf{Rgb-branch.} We use MobileNetV2\cite{MobileNetV2} as our lightweight extractor. We remove the global average pooling layer and fully connected (FC) layer to obtain global features. According to the resolution of the feature maps, the backbone can be divided into 5 stages. 
The output feature maps are denoted as $\mathcal{F}^{R}_{1}$, $\mathcal{F}^{R}_{2},\mathcal{F}^{R}_{3},\mathcal{F}^{R}_{4},\mathcal{F}^{R}_{5}$, and the global feature is denoted as $\mathbf{x}^R$.

\textbf{Seg-branch.} Considering that segmentation label maps are more straightforward and have a higher semantic level than the paired RGB images, we adopt the depth stream in MobileSal\cite{mobilesal} as the feature extractor in seg-branch. It has five stages with same strides and is not as large a capacity as MobileNetV2, denoted MobileNet-L in our paper. 
The output feature maps of five stages are denoted as $\mathcal{F}^{S}_{1},\mathcal{F}^{S}_{2},\mathcal{F}^{S}_{3},\mathcal{F}^{S}_{4},\mathcal{F}^{S}_{5}$, and the global feature is represented as $\mathbf{x}^S$.

\begin{figure}
    \centering
    \includegraphics[width=0.9\linewidth]{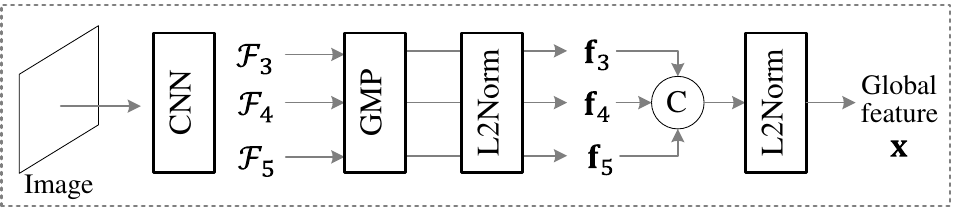}
    \caption{Visualization of Multi-level Concatenation.}
    \label{fig:mc}
\end{figure}

\textbf{Multi-level Concatenation (MC).}
In \fref{fig:mc}, we first apply a global max pooling (GMP) layer to each feature map, $\mathcal{F}_{i}$, to compute single-level features, like 
\begin{equation}
\begin{aligned}
    \mathbf{f}_{i} &= \mathrm{L2Norm}(\mathrm{GMP}(\mathcal{F}_{i})), \\
    % G_{i}^{S} &= \mathrm{L2Norm}(\mathrm{GMP}(\mathcal{F}^{S}_{i})) 
\end{aligned}
\end{equation}
% \in \mathbb{R}^{C_{ri}}, \quad i=\{3,4,5\} 
% \in \mathbb{R}^{C_{si}}, \quad i=\{3,4,5\} 
and concatenate the features from the last 3 layers:
\begin{equation}
\begin{aligned}
    \mathbf{x} &= \mathrm{L2Norm}(\mathrm{Concat}([\mathbf{f}_{3},\mathbf{f}_{4},\mathbf{f}_{5}])), \\
    % G_{S} &= \mathrm{L2Norm}(\mathrm{Concat}([g_{3}^{S},g_{4}^{S},g_{5}^{S}])),
\end{aligned}
\end{equation}
where $\mathbf{x}$ is the unified representation of global features.

\textbf{Triplet VPR Loss.} 
In two training stages, we adopt triplet margin loss\cite{schroff2015facenet} on $(q, p, n)$ as supervision:
\begin{equation}
    \mathcal{L}_{vpr}(\mathbf{x})\!=\!\mathrm{max}(d(\mathbf{x}_q,\mathbf{x}_p)-d(\mathbf{x}_q,\mathbf{x}_n)+m,0), \label{eq:triplet_loss}
\end{equation}
where $\mathbf{x}_q$, $\mathbf{x}_p$, and $\mathbf{x}_n$ refer to global features of query, positive and negative samples. $d(\cdot)$ computes the $L2$ distance of two feature vectors, and margin $m$ is a constant parameter.

\subsection{Group Partition on Training Set}
\label{subsec:group}

As mentioned above, not all samples contain high-quality and helpful teacher knowledge for the student, and even some will hurt the student's performance. So what we need to do is to distill the expected knowledge precisely.
This section addresses this problem simply yet effectively by selecting suitable samples. 

Previous sample-based selective distillation works\cite{selective,liang2021reinforced,ge2018low} rarely consider the student network with specific prior knowledge, resulting in a lack of teacher-student interaction in cross-modal cases.
To avoid this, we let the two pre-trained branches in \sref{subsec:branch} participate in seeking a more accurate partition.

\textbf{Partition Strategy.}
For convenience, ``samples'' mentioned below refer to \textit{sample pairs} (see \sref{subsec:over}).
In this section, we first did some preliminary empirical research, and we found that both pre-trained models have low VPR loss on training sets, but their Recall@$N$ performance cannot achieve 100$\%$ on training sets.
Therefore, the performance of the two branches on a sample, $(q,p)$, can be compared as follows: \textbf{\textit{the ranking of the positive $p$ in the recall list of the query $q$}}.

For convenience, recall rankings of samples of seg-branch and rgb-branch are denoted as $x$ and $y$, respectively.
As shown in \fref{fig:pipeline}, specific definitions for partitioning training sets are as follows:
\vspace{-0.2cm}
$$
\begin{aligned}
{\mathcal{D}_{1}} &=\{(q,p)|x \leq {N_t},y>{N_t}\}, \\
{\mathcal{D}_{2}} &=\{(q,p)|x \leq y \leq {N_t}\}, \\
{\mathcal{D}_{3}} &=\{(q,p)|y<x \leq {N_t}\}, \\
{\mathcal{D}_{4}} &=\{(q,p)|x>{N_t}\},
\end{aligned}
$$
where ${\mathcal{D}_{1}}$ represents the set of samples with $x$ less than or equal to ${N_t}$ and $y$ greater than ${N_t}$, and so on. ${N_t}$ is a constant hyper-parameter.
Intuitively, ${\mathcal{D}_{1}}$ is the most essential and helpful group for knowledge distillation.

\subsection{Weighted Knowledge Distillation}
\label{subsec:kd}
% 额外的0/1和dist-all对比 用于判断哪个群组有效 哪个无效【仅筛选】样本数量和质量的平衡，数量太少效果也不会很好

In order to accurately reflect the impacts of samples, here we weigh the distillation loss based on group partition in \sref{subsec:group}. This way, helpful samples are emphasized, and harmful samples are neglected.

\textbf{Weighting Function.}
Instead of using several constant weights for different groups, we define a function to refine the weights on samples from two perspectives.
One is the knowledge levels of the teacher on each sample; the higher the knowledge level, the greater the weight. 
Another is the knowledge gap between the teacher and the student; the greater the gap, the greater the weight.

Finally, the weight function is defined as:
\begin{equation}\vspace{-0.05cm}
\label{eq:rule}
\varphi(q,p)=\\
\begin{cases}
1+\frac{\min \left(N_{m},y\right)-x}{4 \cdot \ln \left(1+x\right)}, &(q,p) \in {\mathcal{D}_{1}}\\
1+\frac{y-x}{5 \cdot \ln \left(1+x\right)}, &(q,p) \in {\mathcal{D}_{2}}\\
1+\frac{y-x}{4 \cdot \ln \left(1+x\right)}, &(q,p) \in {\mathcal{D}_{3}}\\
0, &(q,p) \in {\mathcal{D}_{4}}
\end{cases}, \vspace{-0.05cm}
\end{equation}
where the non-zero part of $\varphi$ is proportional to $y$ and inversely proportional to $x$. In other words, $\varphi$ is proportional to the performance of the seg-branch on $(q,p)$.
This function makes each sample determine its learning degree.

\textbf{Feature-based Distillation.}
In Stage II, the backbone structure of RGB feature extractor is also MobileNetV2, and the global feature is denoted as $\mathbf{x}^{D}$.
Because $\mathbf{x}^{D}$ and $\mathbf{x}^{S}$ for an image pair $(I^{R}, I^{S})$ are not in the same domain and same dimension, we embed $\mathbf{x}^{D}$ via an additional linear layer in the second-stage training, called transformation function \textit{T}.
We adopt loss\cite{gupta2016cross} in our distillation process:
\begin{equation}\vspace{-0.05cm}
\label{eq:skt}
\mathcal{L}_{kd}(I)=\varphi(q,p) \cdot\left\|\mathbf{x}^{S}_{I}-T\left(\mathbf{x}^{D}_{I}\right)\right\|_{2}^{2},  I \in \{q,p,n\}, \vspace{-0.05cm}
\end{equation}
where $n$ is the negative sample corresponding to $(q,p)$ in the second-stage training. Note that $\mathbf{x}^{S}_{I}$ corresponds to the trained seg-branch in Stage I.
% and $\mathbf{x}^{D}_{I}$ corresponds to the feature extraction in Stage II.

Hence, the extraction model in Stage II can be trained by minimizing the loss as:
\begin{equation} \vspace{-0.05cm}
\mathcal{L}(q, p, n)=\mathcal{L}_{vpr}(\mathbf{x}^{D})+\sum_{\{q, p, n\}}{\mathcal{L}_{kd}(I)}. \vspace{-0.05cm}
\end{equation}

\section{Experiments and Analysis}

\subsection{Datasets and Evaluation}
Datasets with still challenging variations are considered in our work, and the summary is shown in \tref{tbl:dataset}: Mapillary Street Level Sequences (MSLS)\cite{msls}, Pittsburgh\cite{pitts30k}, and Nordland\cite{nordland}.
Compared with MSLS, Pittsburgh contains many urban buildings shot at close range from non-horizontal perspectives, resulting in insufficient categories and instances contained in segmentation images.
That is, MSLS has obvious structural information, while Pittsburgh has relatively weak structural information.
So we train models on these two datasets separately to illustrate that datasets do not limit the effectiveness of StructVPR.
% We defer more details to the Supplementary Material.}

% \textbf{Evaluation Metrics.}
For all datasets, Recall$@N$ is used. A given query is regarded to be correctly localized if at least one of the top $N$ retrieved database images is within a ground truth tolerance of default configuration for these datasets\cite{msls,nord1,pitts30k}.
% small,footnotesize,scriptsize,tiny

\subsection{Implementation Details}
Our method is implemented in PyTorch.
To obtain segmentation images, we use open-source models\footnote{In the main paper, we only show the results of one model, and the rest are in the Supplementary Material, which demonstrates that StructVPR is compatible with many models.}.
All images are resized to $640 \times 480$.
In the SLME module, $C$ is set as 6, including vegetation, dynamic objects, sky, ground, buildings, and other objects, and initial encoded values are 0.5, 0.5, 1, 1, 2, 2.
For {rgb-branch}, we use MobileNetV2 as the backbone, remove the FC layer, and add the MC layer~(\sref{subsec:branch})~(dim=448).
For {seg-branch}, we build a smaller backbone than MobileNetV2 and concatenate the last three levels to conduct global features (dim=480).
For group partition strategy, $N_{t}=10$ and $N_{m}=20$.
\begin{table}
\setlength\tabcolsep{1.6pt}
\renewcommand{\arraystretch}{1.2}
\footnotesize
  \caption{\textbf{Summary of datasets for evaluation.} $++$, $+$, and $-$ indicate degrees from high to low.}
  % 定性评估
  % \textit{Structural information} represents explicit structural information in the data-set's scenarios.
  \vspace{-0.05cm}
  \label{tbl:dataset}
  \centering
  \scalebox{0.9}{\begin{tabular}{l|c|cc|cccc}
  \toprule
  \multirow{2}{*}{Dataset} &{\textbf{Structural}} & \multicolumn{2}{c|}{Environment}  & \multicolumn{4}{c}{Variation}\\
  \cline{3-8}
   &\textbf{information} & \rotatebox{0}{Urban} & \rotatebox{0}{Suburban} &\rotatebox{0}{View} & \rotatebox{0}{Light} & \rotatebox{0}{Long-term} & \rotatebox{0}{Dynamic} \\
  \midrule
MSLS\cite{msls} & $++$ & \checkmark & \checkmark &  $+$  & $+$  & $+$  & $+$\\
Nordland\cite{nordland} & $+$ & & \checkmark& $-$ & $-$ & $++$  & $-$\\
Pittsburgh\cite{pitts30k}& $-$ &\checkmark && $+$ & $-$ & $-$  & $+$ \\
  \bottomrule                  
  \end{tabular}}
\end{table}

\begin{table*}
  \centering
   \caption{Comparison to SOTA methods on datasets. The results of global retrieval and re-ranking with local features are shown.}
  \label{tab:compare_SOTA}
  \scalebox{0.75}{
  \renewcommand{\arraystretch}{1.2}
  \begin{tabular}{ll|c||ccc||ccc||ccc||ccc@{}}
  \toprule
 &\multirow{2}{*}{Method}  &\multirow{2}{*}{Venue}& \multicolumn{3}{c||}{MSLS val} & \multicolumn{3}{c||}{MSLS challenge} & \multicolumn{3}{c||}{Nordland test} & \multicolumn{3}{c}{Pittsburgh30k test}\\% & \multicolumn{3}{c}{Tokyo 24/7}\\
\cline{4-15}
& & & R@1 & R@5 & R@10 & R@1 & R@5 & R@10 & R@1 & R@5 & R@10  & R@1 & R@5 & R@10 \\% & R@1 & R@5 & R@10 \\
\hline
\hline
\multirow{7}{*}{\begin{tabular}[c]{@{}c@{}}Global\\ retrieval\end{tabular}} &
NetVLAD\cite{netvlad} & CVPR'16&53.1 & 66.5 & 71.1 &28.6 & 38.3 & 42.9 & 11.5 & 17.6&  21.9 & 81.9 & 91.2 & 93.7\\% & 57.1 & 72.1 & 77.1 \\
& SFRS\cite{ge2020self} & ECCV'20&58.8 & 68.2 & 71.8 & 30.7 & 39.3 & 42.9 & 14.3 & 23.1 & 27.3 & {71.1} & {81.0} & {84.9}\\% & TD85.4 & 91.1 & 93.3\\
& DELG\cite{delg} & ECCV'20&68.4&78.9 &83.1 & 37.6& 50.5& 54.6 & 27.0&43.3&50.0 & 79.0 & 89.0 & 92.7 \\%& 71.4 & 84.1 & 88.6\\
& Patch-NetVLAD-s\cite{hausler2021patch} & CVPR'21&63.5 & 76.5 & 80.1 & 36.1 & 49.9 & 55.0 &  18.1 & 33.2  & 41.1 & 81.3 & 91.1& 93.4 \\% & 54.6 & 67.6 & 71.4 \\
& Patch-NetVLAD-p\cite{hausler2021patch} & CVPR'21&70.0 & 80.4 & 83.8 & 38.1 & 51.2 & 55.3 & 24.8 & 39.4 & 48.0 & \underline{83.7} & \underline{91.8} & \underline{94.0}\\% & 67.0 & 78.1 & 81.0\\
& TransVPR \cite{transvpr}& CVPR'22&\underline{70.8}&\underline{85.1} & \underline{89.6}  & \underline{48.0} & \underline{67.1} & \underline{73.6} & \underline{31.3} & \underline{53.6} & \underline{64.8} & 73.8 & 88.1 & 91.9\\% & 42.5 & 62.9 & 69.8\\
% \midrule
\cline{2-15}
& \textbf{(A) Ours} & &\textbf{83.0} & \textbf{91.0} & \textbf{92.6} & \textbf{64.5} & \textbf{80.4}& \textbf{83.9} & \textbf{56.1} & \textbf{75.5} & \textbf{82.9} & \textbf{85.1} & \textbf{92.3} & \textbf{94.3}\\% & 63.8 & 73.0 & 78.3\\
\hline
\hline
\multirow{7}{*}{Re-ranking} &
SP-SuperGlue\cite{detone2018superpoint,sarlin2020superglue} & CVPR'20&78.1 & 81.9 & 84.3 & 50.6 & 56.9 & 58.3 & 37.9 & 41.2 & 42.6 & 87.2 & 94.8 & 96.4\\% & 83.2 & 84.4 & 85.4\\
& DELG\cite{delg} & ECCV'20& 83.9 & 89.2 & 90.1 & 56.5 & 65.7 & 68.3 &64.4&70.8&72.7& \underline{89.9} & \underline{95.4} & \underline{96.7} \\%& 85.1 &90.8 & 92.1\\
& Patch-NetVLAD-s\cite{hausler2021patch} & CVPR'21&77.2 & 85.4 & 87.3 & 48.1 & 59.4 & 62.3 & 50.9 & 62.7 & 66.5 & 88.0 & 94.5 & 95.6 \\%& 76.8 & 82.2 & 83.8\\
& Patch-NetVLAD-p\cite{hausler2021patch} & CVPR'21&79.5 & 86.2 & 87.7 & 51.2 & 60.3 & 63.9 & 62.7 & 71.0 & 73.5 & 88.7 & 94.5 & 95.9 \\%& 86.0&88.6& 90.5\\ % 比论文报的结果高
& TransVPR \cite{transvpr}&CVPR'22& {86.8} & 91.2 & 92.4 & 63.9 & 74.0 & 77.5 & \underline{77.8} & \underline{86.8} & 89.3 & 89.0 & 94.9 & 96.2\\% & 77.8 & 81.3 & 93.5\\
% \midrule
\cline{2-15}
& \textbf{(B) Ours-SP-RANSAC} & &\underline{87.3} & \underline{91.4} & \underline{92.8} & \underline{65.5} & \underline{76.3} & \underline{81.3} & 76.8 & 86.3 & \underline{90.1} & 89.4 & 95.2 & 96.5\\% &81.4& 84.2&91.8\\
& \textbf{(B) Ours-SP-SuperGlue} & &\textbf{88.4} & \textbf{94.3} & \textbf{95.0} & \textbf{69.4} & \textbf{81.5} & \textbf{85.6} & \textbf{83.5} & \textbf{93.0} &\textbf{95.0} & \textbf{90.3} & \textbf{96.0} & \textbf{97.3}\\% & 84.7& 87.5&93.4\\
\bottomrule
\end{tabular}}
\end{table*}

\textbf{Training.}
In both training stages, MobileNetV2 is used as the RGB feature extractor and is initialized with the pre-trained weights on ImageNet\cite{krizhevsky2012imagenet},
and MobileNet-L is initialized with random parameters in seg-branch.
In both training stages, we train models on two datasets: Pittsburgh 30k for urban imagery (Pittsburgh dataset) and MSLS for all other conditions. 
The MSLS training set provides GPS coordinates and compass angles, so images most similar to the query field of view are selected as positive samples.
Considering that the Pittsburgh 30k dataset only has location labels, the weakly supervised positive mining strategy proposed in \cite{netvlad} is adopted.
Further details are presented in the Supplementary Material.

\subsection{Comparisons with State-of-the-Art Methods}

\textbf{Baselines.} We compared our method against several state-of-the-art image retrieval-based localization solutions, 
including two methods using global descriptors only:
\textbf{NetVLAD}\cite{netvlad} and \textbf{SFRS}\cite{ge2020self}
, and four methods which additionally perform re-ranking using spatial verification of local features:\textbf{DELG}\cite{delg}, \textbf{Patch-NetVLAD}\cite{hausler2021patch}, \textbf{SP-SuperGlue} and \textbf{TransVPR}\cite{transvpr}.
DELG, Patch-NetVLAD, and TransVPR jointly extract global and local features for image retrieval, while SP-SuperGlue re-ranks NetVLAD retrieved candidates by using SuperGlue\cite{sarlin2020superglue} matcher to match SuperPoint\cite{detone2018superpoint} local features. 
For Patch-NetVLAD, we tested its speed-focused and performance-focused configurations, denoted as Patch-NetVLAD-s and Patch-NetVLAD-p, respectively. More installation details are explained in the Supplementary Material.

\textbf{Re-ranking Backend.}
Candidates obtained through global retrieval are often re-ranked using geometric verification or feature fusion.
In this paper, our main contribution is based on extracting global features, so we use two classic re-ranking methods (i.e., RANSAC and SuperGlue) as the backend to compare with the other two-stage algorithms.
SuperPoint\cite{detone2018superpoint} descriptors are used as our local features.
For SuperGlue, we use the official implementation and configuration.
For RANSAC, the maximum allowed reprojection error of inliers is set to 24.

\textbf{Quantitative comparison.}
\tref{tab:compare_SOTA} compares ours against the retrieval approaches.
There are two settings for StructVPR: (A) global feature similarity search and (B) global retrieval followed by re-ranking with local feature matching (SP-RANSAC, SP-SuperGlue).
In the experiments, we first compare our global model with all methods. Then re-ranking algorithms are used as our backend to compare with those two-stage algorithms.
For all two-stage methods, the top 100 images are further re-ranked. 

In setting (A), ours-global convincingly outperforms all compared global methods by a large margin on MSLS validation, MSLS challenge, and Nordland datasets. Compared with the best global baseline, the absolute increments on Recall$@1$ are 12.2$\%$, 16.5$\%$ and 14.8$\%$, and the ones on Recall$@5$ are 5.9$\%$, 13.3$\%$ and 21.9$\%$ respectively.
More importantly, ours-global is better than almost all two-stage ones on Recall$@5$ and Recall$@10$, which shows the potential of our method for improvement on Recall$@1$.
Ours-global also achieves SOTA results on Pitts30k dataset, with a 1.4$\%$ increment on Recall$@1$ over Patch-NetVLAD.
{We observe that the global model trained on MSLS shows a significant improvement compared with previous SOTA methods, while the one by Pitts is relatively mediocre. Such results are due to differences in datasets, that is, whether the scene contains enough structural information.
This experimental phenomenon also proves that datasets do not limit StructVPR: on datasets with obvious structural information, it can effectively improve performance; on datasets with insignificant structural information, it can achieve effective distillation without hurting the RGB performance.}

In setting (B), ours-SP-SuperGlue achieves SOTA performance on all datasets, illustrating the importance of global retrieval compared with SP-SuperGlue.
Ours-SP-SuperGlue outperformed the best baseline (TransVPR) on key benchmarks by an average of 3.53\% (absolute R@$1$ increase) and by 4.48\% (absolute R$@5$ increase).

% 有些ransac完效果反而变差 说明不如我们全局本身可以提供的信息；我们加的后端不是完全系统适配的，所以有的表现好 有的还不如我们

\begin{table}
  \centering
\caption{\textbf{Latency and memory footprint.} The following data is measured on NVIDIA GeForce RTX 3090 GPU and Intel Xeon Gold 6226R CPU. For global retrieval methods, matching time and memory requirements are negligible.}
\scalebox{0.7}{
\begin{tabular}{cl|c|c|c}
\toprule
& Method & \begin{tabular}[c]{@{}c@{}}Extraction\\ latency (ms)\end{tabular} &         \begin{tabular}[c]{@{}c@{}}Matching\\ time (s)\end{tabular} & \begin{tabular}[c]{@{}c@{}}Memory \\ (MB)\end{tabular} \\
\midrule
\multirow{3}{*}{\begin{tabular}[c]{@{}c@{}}Global\\ retrieval\end{tabular}}
& NetVLAD\cite{netvlad} & 40 & $-$ & $-$ \\
& SFRS\cite{ge2020self} & 207 & $-$ & $-$ \\
& Ours & 2.25 & $-$ & $-$\\
\midrule
\multirow{5}{*}{Re-ranking}
& SP-SuperGlue\cite{detone2018superpoint,sarlin2020superglue} & 35 & 6.4 & 1.93 \\
& DELG\cite{delg} & 199 & 45.7 & 0.37 \\
& Patch-NetVLAD-s\cite{hausler2021patch} & 42.5 & 3.29 & 1.82 \\
& Patch-NetVLAD-p\cite{hausler2021patch} & 625 &  25.6 & 44.14 \\
& TransVPR & 8 & 3.5 & 1.17 \\
% & Ours-SP-SuperGlue & 35 &  &\\
\bottomrule
\end{tabular}
}
\label{tab:latency_memory}
\end{table}

Please refer to Supplementary Material for detailed qualitative results.

\textbf{Latency and memory footprint.}
In real-world VPR systems, latency and resource consumption are important factors.
As shown in \tref{tab:latency_memory}, latency and memory requirements refer to processing a single query image.

Ours, with only the global retrieval step, has a great advantage in latency over all other algorithms.
Ours-SP-SuperGlue achieves SOTA recall performance, and the extra computational latency and memory are the same as SP-SuperGlue. It is 5.3 times and 16.8 times faster than DELG and Patch-NetVLAD-p in feature extraction, and it is 7.1 times and 4 times faster than them in spatial matching. 

In summary, previous high-accuracy two-stage methods mainly rely on re-ranking, which comes at a high computational cost. On the contrary, our global retrieval results are good with low computational cost, achieving a better balance of accuracy and computation.

% reranking的成本代价；在哪个数据集上

\subsection{Ablation Studies}
\label{sec:ablation}
We conduct several ablation experiments to validate the proposed modules in our work. The results on MSLS are given here, and the results on other datasets are presented in the Supplementary Material.

\begin{figure}
    \centering
    \includegraphics[width=0.95\linewidth]{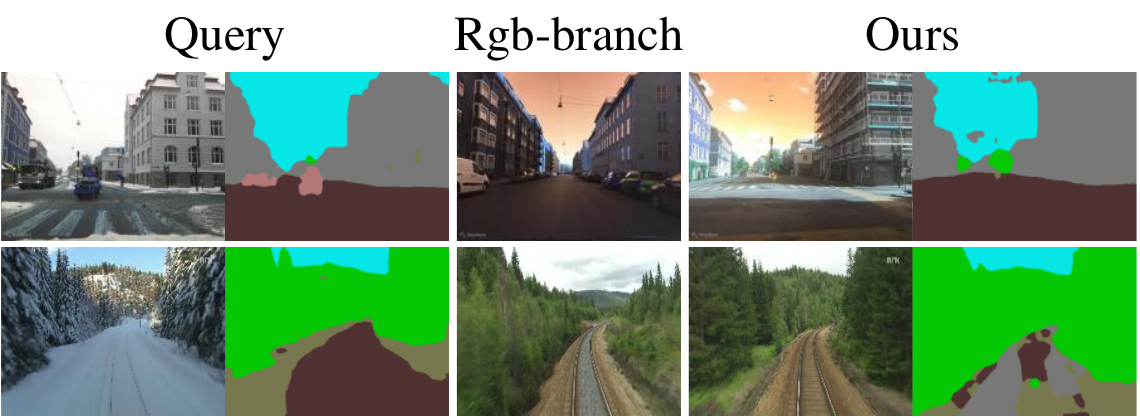}
    \caption{\textbf{Qualitative results.} In these examples, StructVPR successfully retrieves matching reference images, while rgb-branch produces incorrect place matches.}
    \label{fig:plot}
\end{figure}

\begin{table}
  \centering
   \caption{Comparison to other fusion solutions on MSLS. Extraction latency includes the time to generate segmentation images.}
  \label{tab:compare_arch}
  \scalebox{0.7}{
  \renewcommand{\arraystretch}{1.2}
  \begin{tabular}{l||ccc||ccc||c}
  \toprule
\multirow{2}{*}{Arch}   & \multicolumn{3}{c||}{MSLS val} & \multicolumn{3}{c||}{MSLS challenge} & Extraction\\
\cline{2-7}
& R@1 & R@5 & R@10 & R@1 & R@5 & R@10 & latency (ms) \\
\midrule
SEG & 67.7 & 80.0 & 83.1 & 43.4 & 58.9 & 65.8 & 4$+${385}\\
RGB & 75.8 & 85.3 & 87.3 & 55.1 & 71.9 & 76.4 & 2.25\\
Multi-task & 75.3 & 86.2 & 88.7 & 56.8 & 72.3& 77.1 & 2.25\\
C-feat & 75.5 & 86.5 & 88.8 & 55.9 & 73.3 & 78.1 & 6.25$+$385\\
C-input & \underline{77.7} & \underline{88.4} & \underline{91.4} & \underline{59.7} & \underline{76.9} & \underline{81.2} & 21.67$+$385\\
\midrule
KD(Ours) & \textbf{83.0} & \textbf{91.0} & \textbf{92.6} & \textbf{64.5} & \textbf{80.4} & \textbf{83.9} & 2.25\\
% KD是最终的
% seg是加权的
% 单张图像的处理 process a single query image
\bottomrule
\end{tabular}}
\end{table}

\textbf{Fusion Solutions.}
In the preliminary empirical studies, it has been demonstrated that segmentation images benefit VPR, and many solutions can achieve the fusion of two modalities in global retrieval. 
Here we give several feasible solutions and compare them:
\begin{itemize}
    \item Concat-input: Concatenate RGB image and encoded segmentation label map in channel $C$ as input and train the model. Segmentation is required during testing. % 网络变大、训练很慢、输入处理 dim=448 encoded
    \item Concat-feat: Two feature vectors from two separate models are concatenated into a global feature. Segmentation is required during testing. % 是否需要协同训练？还是基于原有的两个分支就可以？
    \item Multi-task: Use the encoder-decoder structure, share the encoder with two tasks, and use the segmentation map as the supervision of the decoder. Segmentation is not required during testing. %交叉熵损失 one-hot概率分布 VS 0/1 dim=448
    \item Knowledge distillation (Ours): 
    % Apply knowledge distillation from SEG to RGB. The final loss is the sum of the VPR loss and the weighted feature-based distillation loss. 
    Segmentation is not required during testing.% on \textbf{} samples
\end{itemize}

% \begin{figure}
%     \centering
%     \includegraphics[width=0.9\linewidth]{fig/RN_mslsval.pdf}
%     \caption{\textbf{Comparison with state-of-the-art on MSLS val. set.} We show the Recall$@N$ performance of Ours compared with other methods. Results w/o re-ranking are depicted in dotted line, while results with re-ranking are depicted in solid line.}
%     \label{fig:msls_compare}
% \end{figure}

We train and compare them on MSLS and count the extraction latency during testing. Extraction latency is the time to obtain global features from a single RGB image.

As shown in \tref{tab:compare_arch}, all fusion algorithms perform better than the two separate branches (RGB, SEG) on the MSLS dataset.
But the performance of Concat-feat is limited, which may be due to a lack of coherence between two modalities.
Although both Concat-input and KD perform relatively well, Concat-input requires additional processing of segmentation images during testing and has a larger model compared with KD. % 说明KD效果最好？
% 除个别R@1的数据外 基本满足M<CF<CI<KD
Neither Multi-task nor KD use segmentation images during testing. However, it is worth noting that the training of Multi-task has high requirements for parameter tuning, and KD is more direct and interpretable than Multi-task.
Here we qualitatively demonstrate the advantages that StructVPR brings in \fref{fig:plot}.

\textbf{Segmentation label map encoding.}
% 只针对seg的预先训练分支
% 3类且加权的呢？
% 3类3值不同于6类3值
The number of original semantic labels of the open-source model is 150.
In \sref{subsec:encoding_module}, we use an encoding function to convert segmentation images into weighted one-hot label maps.
Considering that $C$ and the prior weight setting are the most important hyper-parameters for our work, we perform ablation experiments on them.
We also test $C=3,150$ in {seg-branch} for the number of clustered classes. After clustering, the redefined 3 categories are \{sky\&ground, dynamic objects, and static objects\}.
In addition, we present the results of different prior weight settings for $C=6$.

\begin{figure}
    \centering
    \captionsetup[subfloat]{labelsep=none,format=plain,labelformat=empty}
    \subfloat[]{
    %   \label{fig:seg_val}
      \includegraphics[width=0.48\linewidth]{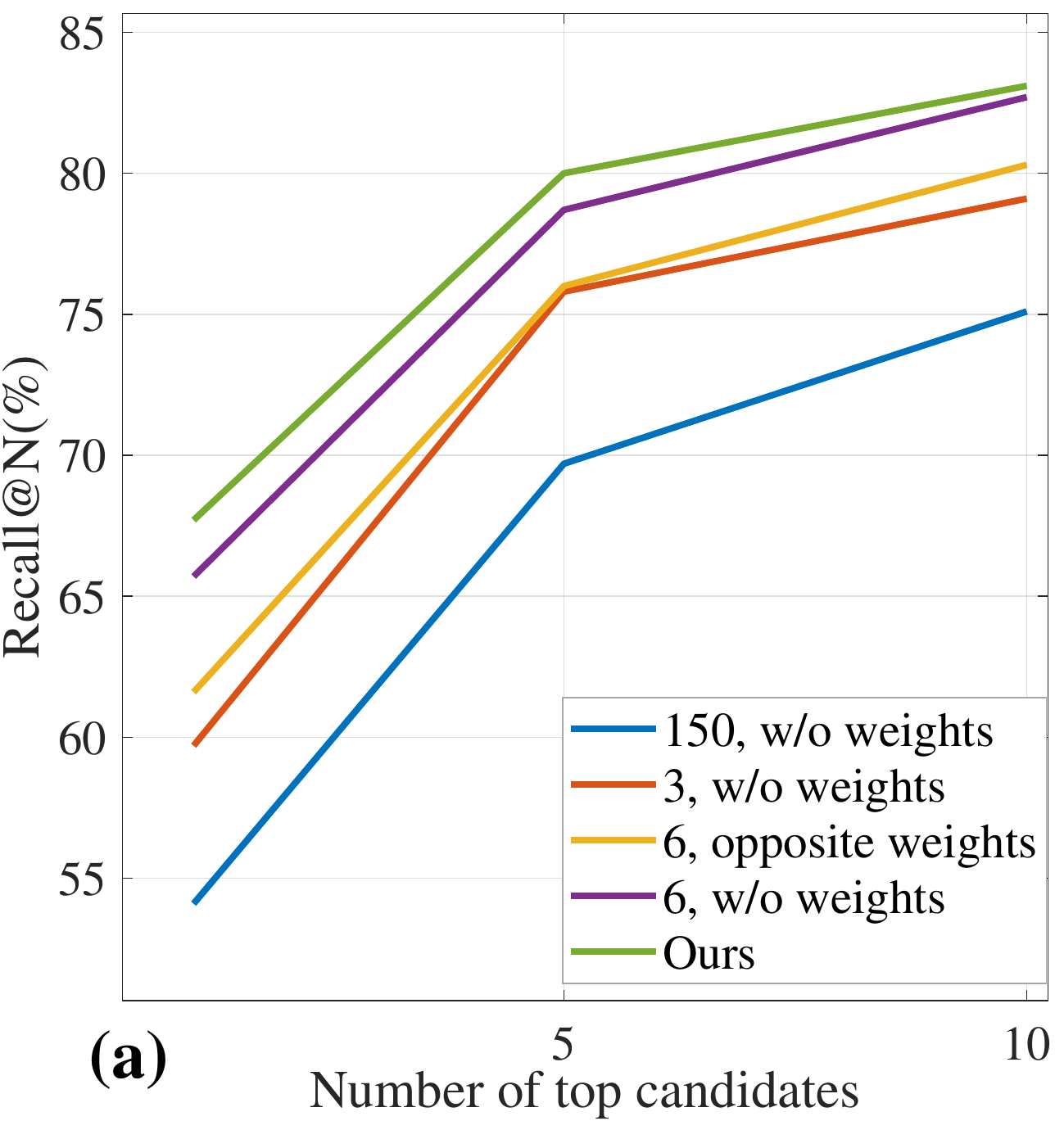}}
    \subfloat[]{
    %   \label{fig:seg_test}
      \includegraphics[width=0.48\linewidth]{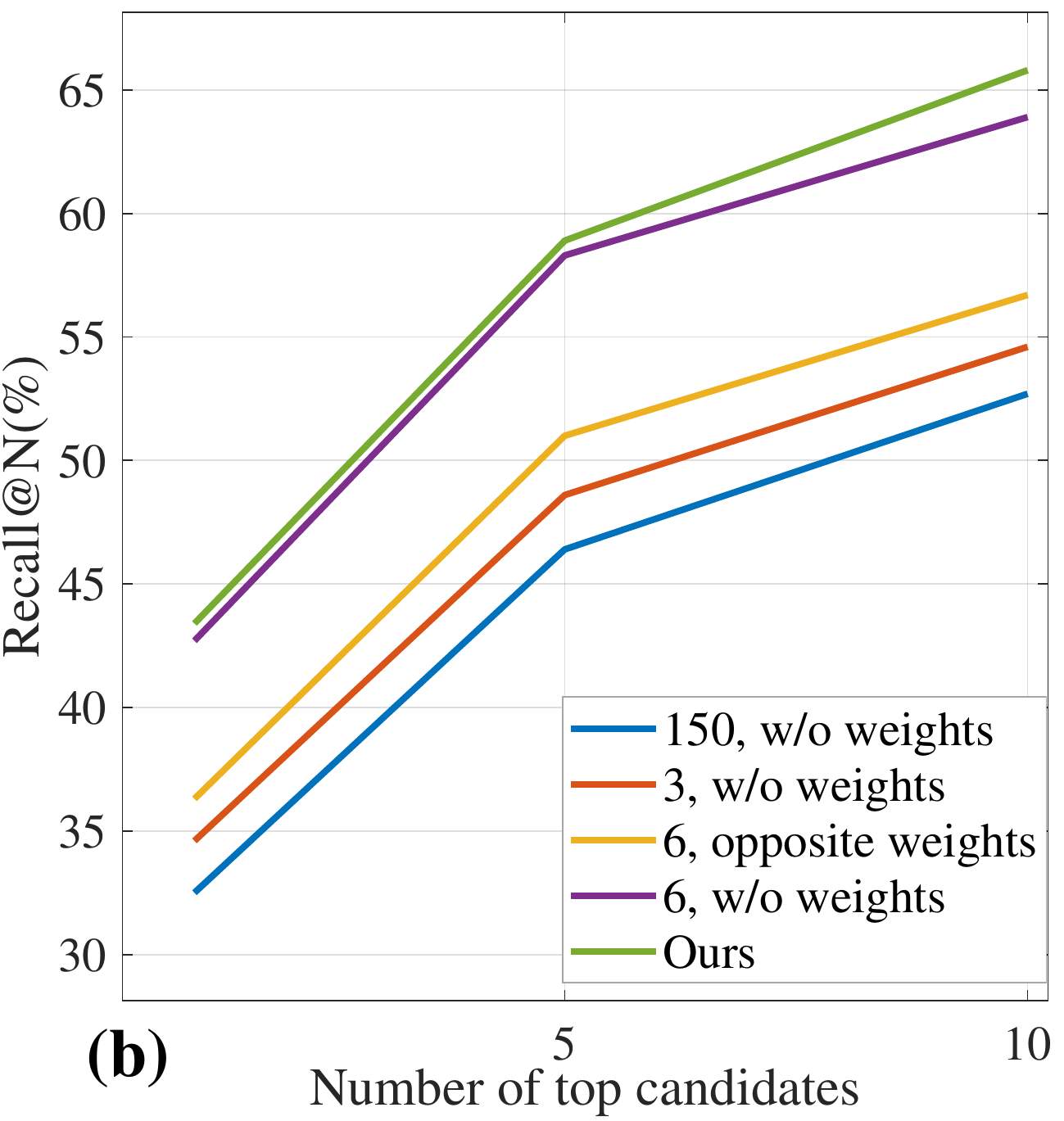}}
      \vspace{-0.5cm}
    \caption{Performance of {seg-branch} with different number of clustered classes on (a) MSLS val. and (b) MSLS test dataset.}
    \label{fig:class}
\end{figure}

% 表2的SEG  引导收敛的作用
% notes memory和推理时长的区别是显而易见的 可能就是具体数值上 所以没啥必要 supp都不用？
% 150相当于没有这个聚类的过程
% seg的时间 都是离线的 所以没必要讲

% \tref{tab:number_abl} 
\fref{fig:class} shows the results on the MSLS dataset.
It can be seen that a too-small value of $C$ is not advisable, which will lose the uniqueness of the scene in our view, and the final performance is poor when $C=150$, which means that too fine-grained segmentation makes training more difficult.
Finally, $C$ is set as 6.
$\dagger$ stands for the opposite weight configuration: 0.5 for static objects and other objects, 2 for dynamic objects and vegetation.
Comparing the results of unweighted and two weighted cases, we can qualitatively conclude that introducing appropriate prior weights can guide the model to converge better.
Further analysis on prior weights is presented in the Supplementary Material.

\textbf{Group partition strategy and the impact of groups.}
% 体现选择很重要 以及我们的划分方式的优势
% 说明GP-D和GP-S在选择层面是一致的 比-S好的
% 下一部分说明函数设计上带来的区别
This paper proposes that not all samples are helpful in distillation and selects samples by group partitioning. At the same time, we propose using two pre-trained branches to participate in group partition together, instead of just using a separate one, to improve the accuracy of group partition and further achieve more accurate weighted KD.
In this experiment, we perform selective KD to verify the importance of each group, which is a degenerate version of weighting for samples. In our expectation, samples of different groups should have different effects on distillation.

\begin{figure}[t]
  \centering
   \includegraphics[width=0.95\linewidth]{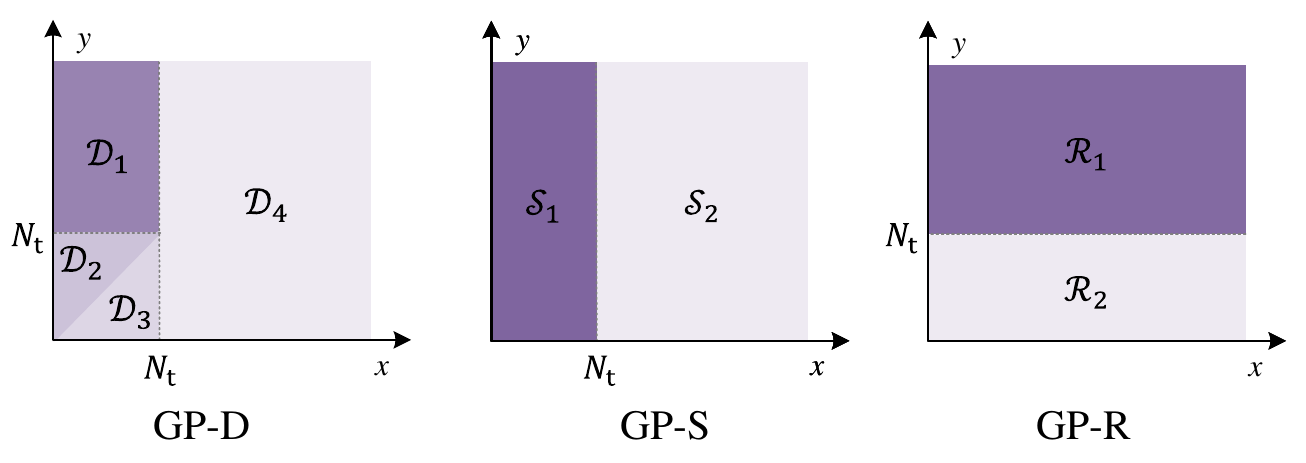}
   \vspace{-0.3cm}
   \caption{\textbf{Visualization of different group partition strategies.} Compared with our strategy (GP-D), which uses two models, both GP-S and GP-R have only one model involved in the partition.}
   \label{fig:group}
\end{figure}

We compare our standard partition strategy (GP-D) with two degenerate configurations, as shown in \fref{fig:group}:
\begin{itemize}
    \item \textit{Group partition with seg-branch}  (GP-S). Only the teacher network participates in the group partition.
    \item \textit{Group partition with rgb-branch} (GP-R). Only the student network participates in the group partition.
\end{itemize}

\begin{table}
  \centering
\caption{Performance of selective distillation with different samples. None refers to the {rgb-branch} without distillation and All refers to non-selective distillation.}
\vspace{-0.3cm}
\label{tab:group_abl}
  \scalebox{0.7}{
 \renewcommand{\arraystretch}{1.2}
\begin{tabular}{c|c|ccc|ccc}
\toprule
\multirow{2}{*}{\begin{tabular}[c]{@{}c@{}}Group for \\ distillation\end{tabular}} & \multirow{2}{*}{\begin{tabular}[c]{@{}c@{}} Sample\\ratio \end{tabular}}&\multicolumn{3}{c|}{MSLS val} & \multicolumn{3}{c}{MSLS challenge} \\
\cline{3-8}
& & R@1      & R@5     & R@10    & R@1       & R@5       & R@10      \\
\midrule
None & 0\% & 75.8 & 85.3& 87.3 & 55.1 & 71.9 & 76.4\\
\textbf{All} & 100\% &78.4 & 87.4& 90.1 & 59.1 & 73.5& 79.3\\ % 0707
\midrule
${\mathcal{D}_{1}}$ & 4.11\% & 78.8 & \underline{89.2}& 91.1 & \underline{62.3} & 76.8 & 80.9\\ % 0803
${\mathcal{D}_{2}}$ & 50.67\%  & 80.0 & 88.8& 90.6 & 59.1&75.5&79.3\\ % 0806
${\mathcal{D}_{3}}$ & 14.77\%  &77.7 & 87.4 & 89.2 & 57.8&74.4&79.1 \\ % 0524
${\mathcal{S}_{1}}$& 69.55\% & \underline{81.6} & 88.9 & \underline{91.2} & {62.2} & \underline{77.2} & \underline{82.0}\\ % 0813
${\mathcal{S}_{2}}({\mathcal{D}_{4}})$ & 30.45\% & 71.6&83.7&85.1 & 48.2 & 63.9 & 68.7 \\ % 0518
${\mathcal{R}_{1}}$ & 73.24\% &79.3&87.6&88.9 & 57.0&74.1&78.2\\ % 0806
${\mathcal{R}_{2}}$ & 26.76\% &73.8&82.7&85.6 & 51.0&67.0&72.3\\ % 0518
\midrule
\textbf{Ours} & 69.55\% & \textbf{83.0} & \textbf{91.0} & \textbf{92.6} & \textbf{64.5} & \textbf{80.4}& \textbf{83.9}\\
\bottomrule
\end{tabular}}
\end{table}

For selective KD, GP-D uses samples belonging to $\{{\mathcal{D}_{1}}, {\mathcal{D}_{2}}, {\mathcal{D}_{3}}\}$, GP-R uses $\{{\mathcal{S}_{1}}\}$, and GP-S uses $\{{\mathcal{R}_{1}}\}$. The meaning of ${\mathcal{S}_{1}},{\mathcal{S}_{1}},{\mathcal{S}_{1}},{\mathcal{S}_{1}}$ can refer to \sref{subsec:group}.

As shown in \tref{tab:group_abl}, we select a single group, in turn, to participate in distillation and evaluate its effectiveness. 
The results show that ${\mathcal{S}_{2}}$ and ${\mathcal{R}_{2}}$ harm distillation and other groups have different degrees of positive impact.
\textbf{All} stands for non-selective KD using all samples in KD. However, its results are not as good as ${\mathcal{S}_{1}}$ and ${\mathcal{D}_{1}}$, indicating that the quality of samples is as important as the quantity for KD.
So it is necessary to select suitable samples in distillation.

In this experiment, GP-S performs better than GP-R, and GP-D degenerates to GP-S.
Furthermore, our partitioning strategy will show a considerable advantage over GP-S in weighted knowledge distillation.

\textbf{Weighting distillation loss.}
% 一个是连续函数的好处，一个是进一步说明划分的好处
% 区别是两个变量和群内权重函数斜率设计
We compare our weighting function (\eref{eq:rule}) with constant weights to illustrate the importance of fine-grained weight assignment for samples.
We have verified the importance of the groups above, which guides the design \eref{eq:rule}. The more essential samples for distillation, the higher the weight.

To further illustrate the advantages of GP-D, we also compare GP-D and GP-S with weighting functions.
For GP-S, \eref{eq:rule} degenerates into the following form:
\begin{equation}
\label{eq:rule1}
\varphi'(q,p)=\\
\begin{cases}
1+\frac{1}{4 \cdot \ln \left(1+RN_{s}\right)}, &(q,p) \in {\mathcal{S}_{1}}\\
0, &(q,p) \in {\mathcal{S}_{2}}
\end{cases},
\end{equation}
where the performance gap between the student network and the teacher network cannot be reflected on samples.

\begin{table}
  \centering
\caption{Performance of weighted distillation with different weights. The weights correspond to $\mathcal{S}_{1}$-$\mathcal{S}_{2}$ and $\mathcal{D}_{1}$-$\mathcal{D}_{2}$-$\mathcal{D}_{3}$-$\mathcal{D}_{4}$, and GP-S(1-0) is equal to GP-D(1-1-1-0).}
\vspace{-0.3cm}
\label{tab:weight_abl}
  \scalebox{0.7}{
 \renewcommand{\arraystretch}{1.2}
\begin{tabular}{c|ccc|ccc}
\toprule
\multirow{2}{*}{\begin{tabular}[c]{@{}c@{}}Weight\end{tabular}} & \multicolumn{3}{c|}{MSLS val} & \multicolumn{3}{c}{MSLS challenge} \\
\cline{2-7}
& R@1      & R@5     & R@10    & R@1       & R@5       & R@10      \\
\midrule
GP-S(1-0) &  {81.6} & 88.9 & 91.2 & {62.2} & 77.2 & 82.0\\
GP-D(8-4-1-0) & \underline{82.2} & 88.9 & 91.5 & {62.3} &\underline{78.9} & 82.4\\ % 0719
\midrule
GP-S($\varphi'(q,p)$)& 79.7 & \underline{89.5} & \underline{92.0} & \underline{62.7} & {78.4} & \underline{83.7}\\ % 0726
GP-D($\varphi(q,p)$) & \textbf{83.0} & \textbf{91.0} & \textbf{92.6} & \textbf{64.5} & \textbf{80.4} & \textbf{83.9} \\
\bottomrule
\end{tabular}}
\end{table}

In \tref{tab:weight_abl}, GP-D(8-4-1-0) performs better than GP-S(1-0), and GP-D($\varphi(q,p)$) is better than GP-S($\varphi'(q,p)$), which illustrate the importance of precise partition.
Moreover, for both GP-D and GP-S, the weight function performs better than the discrete constant weights, respectively, showing the advantages of setting weights for each sample.
% 无法体现两者之间的差距 不如两个变量；也无法区分内部；
% 无论是针对群的定值加权还是样本，加权后的GP-S不如GP-D。
% 简单的选择+合适的整体加权、针对group的加权、针对sample的加权
% 通过划分和函数形式的表达，来对每个样本实现一种【近似的】逐个分析
\vspace{-0.5cm}
\section{Discussion}
\label{sec:discussion}
\textbf{Conclusion.} In this paper, we address the problem of using segmentation information to enhance the structural knowledge in RGB global representations, attempting to replace the re-ranking process.
To avoid the computation of segmentation during testing, we use the framework of knowledge distillation.
We find that the teacher's knowledge is not the more, the better; samples have different impacts on knowledge distillation, and some are even harmful.
So we propose a weighted knowledge distillation method to partition samples and weigh the distillation loss for each sample.
Experimental results show that StructVPR achieves SOTA performance among methods with only RGB global retrieval.
Compared with two-stage methods, it is competitive with a better balance of accuracy and computation.

\textbf{Limitations and Future Work.}
% 自觉不足
Nevertheless, some challenges remain.
{The clustered number and encoding values of  semantic labels are chosen from five settings (\fref{fig:class}), and it calls for further advancements (details in appendix).}
Considering the robustness of StructVPR to SLME, smaller SEG models can be used to reduce costs.
Furthermore, we believe that StructVPR is applicable to other tasks or modalities as long as multi-modal information is complementary at the sample level for tasks.

{\small
\bibliographystyle{ieee_fullname}
    \bibliography{ref}

\begin{thebibliography}{10}\itemsep=-1pt

\bibitem{netvlad}
Relja Arandjelovic, Petr Gronat, Akihiko Torii, Tomas Pajdla, and Josef Sivic.
\newblock {NetVLAD}: Cnn architecture for weakly supervised place recognition.
\newblock In {\em CVPR}, pages 5297--5307, 2016.

\bibitem{asami2017domain}
Taichi Asami, Ryo Masumura, Yoshikazu Yamaguchi, Hirokazu Masataki, and Yushi
  Aono.
\newblock Domain adaptation of dnn acoustic models using knowledge
  distillation.
\newblock In {\em ICASSP}, pages 5185--5189, 2017.

\bibitem{wasabi}
Assia Benbihi, St{\'e}phanie Arravechia, Matthieu Geist, and C{\'e}dric
  Pradalier.
\newblock Image-based place recognition on bucolic environment across seasons
  from semantic edge description.
\newblock In {\em IEEE Int. Conf. Robot. Autom.}, pages 3032--3038, 2020.

\bibitem{bucilu2006model}
Cristian Buciluǎ, Rich Caruana, and Alexandru Niculescu-Mizil.
\newblock Model compression.
\newblock In {\em ACM SIGKDD}, pages 535--541, 2006.

\bibitem{camara2020visual}
Luis~G Camara and Libor P{\v{r}}eu{\v{c}}il.
\newblock Visual place recognition by spatial matching of high-level cnn
  features.
\newblock {\em Robot. Autom. Syst.}, 133:103625, 2020.

\bibitem{delg}
Bingyi Cao, Andre Araujo, and Jack Sim.
\newblock Unifying deep local and global features for image search.
\newblock In {\em ECCV}, pages 726--743, 2020.

\bibitem{chen2017learning}
Guobin Chen, Wongun Choi, Xiang Yu, Tony Han, and Manmohan Chandraker.
\newblock Learning efficient object detection models with knowledge
  distillation.
\newblock {\em NeurIPS}, 30, 2017.

\bibitem{chen2017deep}
Zetao Chen, Adam Jacobson, Niko S{\"u}nderhauf, Ben Upcroft, Lingqiao Liu,
  Chunhua Shen, Ian Reid, and Michael Milford.
\newblock Deep learning features at scale for visual place recognition.
\newblock In {\em IEEE Int. Conf. Robot. Autom.}, pages 3223--3230, 2017.

\bibitem{dai2021learning}
Rui Dai, Srijan Das, and Fran{\c{c}}ois Bremond.
\newblock Learning an augmented rgb representation with cross-modal knowledge
  distillation for action detection.
\newblock In {\em ICCV}, pages 13053--13064, 2021.

\bibitem{detone2018superpoint}
Daniel DeTone, Tomasz Malisiewicz, and Andrew Rabinovich.
\newblock Superpoint: Self-supervised interest point detection and description.
\newblock In {\em CVPR}, pages 224--236, 2018.

\bibitem{dusmanu2019d2}
Mihai Dusmanu, Ignacio Rocco, Tomas Pajdla, Marc Pollefeys, Josef Sivic,
  Akihiko Torii, and Torsten Sattler.
\newblock D2-net: A trainable cnn for joint description and detection of local
  features.
\newblock In {\em CVPR}, pages 8092--8101, 2019.

\bibitem{garcia2018modality}
Nuno~C Garcia, Pietro Morerio, and Vittorio Murino.
\newblock Modality distillation with multiple stream networks for action
  recognition.
\newblock In {\em ECCV}, pages 103--118, 2018.

\bibitem{xview}
Abel Gawel, Carlo Del~Don, Roland Siegwart, Juan Nieto, and Cesar Cadena.
\newblock X-view: Graph-based semantic multi-view localization.
\newblock {\em Robot. Autom. lett.}, 3(3):1687--1694, 2018.

\bibitem{ge2018low}
Shiming Ge, Shengwei Zhao, Chenyu Li, and Jia Li.
\newblock Low-resolution face recognition in the wild via selective knowledge
  distillation.
\newblock {\em IEEE TIP}, 28(4):2051--2062, 2018.

\bibitem{ge2020self}
Yixiao Ge, Haibo Wang, Feng Zhu, Rui Zhao, and Hongsheng Li.
\newblock Self-supervising fine-grained region similarities for large-scale
  image localization.
\newblock In {\em ECCV}, pages 369--386, 2020.

\bibitem{gordo2017end}
Albert Gordo, Jon Almazan, Jerome Revaud, and Diane Larlus.
\newblock End-to-end learning of deep visual representations for image
  retrieval.
\newblock {\em IJCV}, 124(2):237--254, 2017.

\bibitem{gou2021knowledge}
Jianping Gou, Baosheng Yu, Stephen~J Maybank, and Dacheng Tao.
\newblock Knowledge distillation: A survey.
\newblock {\em IJCV}, 129(6):1789--1819, 2021.

\bibitem{gupta2016cross}
Saurabh Gupta, Judy Hoffman, and Jitendra Malik.
\newblock Cross modal distillation for supervision transfer.
\newblock In {\em CVPR}, pages 2827--2836, 2016.

\bibitem{hao2020labelenc}
Miao Hao, Yitao Liu, Xiangyu Zhang, and Jian Sun.
\newblock Labelenc: A new intermediate supervision method for object detection.
\newblock In {\em ECCV}, pages 529--545, 2020.

\bibitem{hausler2021patch}
Stephen Hausler, Sourav Garg, Ming Xu, Michael Milford, and Tobias Fischer.
\newblock {Patch-NetVLAD}: Multi-scale fusion of locally-global descriptors for
  place recognition.
\newblock In {\em CVPR}, pages 14141--14152, 2021.

\bibitem{hinton2015distilling}
Geoffrey Hinton, Oriol Vinyals, Jeff Dean, et~al.
\newblock Distilling the knowledge in a neural network.
\newblock {\em arXiv preprint arXiv:1503.02531}, 2(7), 2015.

\bibitem{hoffman2016learning}
Judy Hoffman, Saurabh Gupta, and Trevor Darrell.
\newblock Learning with side information through modality hallucination.
\newblock In {\em CVPR}, pages 826--834, 2016.

\bibitem{dasgil}
Hanjiang Hu, Zhijian Qiao, Ming Cheng, Zhe Liu, and Hesheng Wang.
\newblock Dasgil: Domain adaptation for semantic and geometric-aware
  image-based localization.
\newblock {\em IEEE TIP}, 30:1342--1353, 2020.

\bibitem{khaliq2019holistic}
Ahmad Khaliq, Shoaib Ehsan, Zetao Chen, Michael Milford, and Klaus
  McDonald-Maier.
\newblock A holistic visual place recognition approach using lightweight cnns
  for significant viewpoint and appearance changes.
\newblock {\em IEEE Trans. Robot.}, 36(2):561--569, 2019.

\bibitem{krizhevsky2012imagenet}
Alex Krizhevsky, Ilya Sutskever, and Geoffrey~E Hinton.
\newblock Imagenet classification with deep convolutional neural networks.
\newblock {\em NeurIPS}, 25:1097--1105, 2012.

\bibitem{li2017large}
Jinyu Li, Michael~L Seltzer, Xi Wang, Rui Zhao, and Yifan Gong.
\newblock Large-scale domain adaptation via teacher-student learning.
\newblock {\em arXiv preprint arXiv:1708.05466}, 2017.

\bibitem{liang2021reinforced}
Shining Liang, Ming Gong, Jian Pei, Linjun Shou, Wanli Zuo, Xianglin Zuo, and
  Daxin Jiang.
\newblock Reinforced iterative knowledge distillation for cross-lingual named
  entity recognition.
\newblock In {\em ACM SIGKDD}, pages 3231--3239, 2021.

\bibitem{lowry2015visual}
Stephanie Lowry, Niko S{\"u}nderhauf, Paul Newman, John~J Leonard, David Cox,
  Peter Corke, and Michael~J Milford.
\newblock Visual place recognition: A survey.
\newblock {\em IEEE Trans. Robot.}, 32(1):1--19, 2015.

\bibitem{sta}
Feng Lu, Baifan Chen, Xiang-Dong Zhou, and Dezhen Song.
\newblock {STA-VPR}: Spatio-temporal alignment for visual place recognition.
\newblock {\em Robot. Autom. lett.}, 6(3):4297--4304, 2021.

\bibitem{masone2021survey}
Carlo Masone and Barbara Caputo.
\newblock A survey on deep visual place recognition.
\newblock {\em IEEE Access}, 9:19516--19547, 2021.

\bibitem{mohedano2016bags}
Eva Mohedano, Kevin McGuinness, Noel~E O'Connor, Amaia Salvador, Ferran
  Marques, and Xavier Gir{\'o}-i Nieto.
\newblock Bags of local convolutional features for scalable instance search.
\newblock In {\em ACM Int. Conf. Multimedia Retr}, pages 327--331, 2016.

\bibitem{noh2017large}
Hyeonwoo Noh, Andre Araujo, Jack Sim, Tobias Weyand, and Bohyung Han.
\newblock Large-scale image retrieval with attentive deep local features.
\newblock In {\em ICCV}, pages 3456--3465, 2017.

\bibitem{nord1}
Daniel Olid, Jos{\'e}~M F{\'a}cil, and Javier Civera.
\newblock Single-view place recognition under seasonal changes.
\newblock {\em arXiv preprint arXiv:1808.06516}, 2018.

\bibitem{paolicelli2022learning}
Valerio Paolicelli, Antonio Tavera, Carlo Masone, Gabriele Berton, and Barbara
  Caputo.
\newblock Learning semantics for visual place recognition through multi-scale
  attention.
\newblock In {\em Int. Conf. Image Anal. Process.}, pages 454--466, 2022.

\bibitem{peng2021semantic}
Guohao Peng, Yufeng Yue, Jun Zhang, Zhenyu Wu, Xiaoyu Tang, and Danwei Wang.
\newblock Semantic reinforced attention learning for visual place recognition.
\newblock In {\em IEEE Int. Conf. Robot. Autom.}, pages 13415--13422, 2021.

\bibitem{piasco2019learning}
Nathan Piasco, D{\'e}sir{\'e} Sidib{\'e}, Val{\'e}rie Gouet-Brunet, and
  C{\'e}dric Demonceaux.
\newblock Learning scene geometry for visual localization in challenging
  conditions.
\newblock In {\em IEEE Int. Conf. Robot. Autom.}, pages 9094--9100, 2019.

\bibitem{piasco2021improving}
Nathan Piasco, D{\'e}sir{\'e} Sidib{\'e}, Val{\'e}rie Gouet-Brunet, and
  C{\'e}dric Demonceaux.
\newblock Improving image description with auxiliary modality for visual
  localization in challenging conditions.
\newblock {\em IJCV}, 129(1):185--202, 2021.

\bibitem{radenovic2018fine}
Filip Radenovi{\'c}, Giorgos Tolias, and Ond{\v{r}}ej Chum.
\newblock Fine-tuning cnn image retrieval with no human annotation.
\newblock {\em IEEE TPAMI}, 41(7):1655--1668, 2018.

\bibitem{ren2021learning}
Sucheng Ren, Yong Du, Jianming Lv, Guoqiang Han, and Shengfeng He.
\newblock Learning from the master: Distilling cross-modal advanced knowledge
  for lip reading.
\newblock In {\em CVPR}, pages 13325--13333, 2021.

\bibitem{revaud2019learning}
Jerome Revaud, Jon Almaz{\'a}n, Rafael~S Rezende, and Cesar Roberto~de Souza.
\newblock Learning with average precision: Training image retrieval with a
  listwise loss.
\newblock In {\em ICCV}, pages 5107--5116, 2019.

\bibitem{MobileNetV2}
Mark Sandler, Andrew Howard, Menglong Zhu, Andrey Zhmoginov, and Liang-Chieh
  Chen.
\newblock Mobilenetv2: Inverted residuals and linear bottlenecks.
\newblock In {\em CVPR}, pages 4510--4520, 2018.

\bibitem{sarlin2019coarse}
Paul-Edouard Sarlin, Cesar Cadena, Roland Siegwart, and Marcin Dymczyk.
\newblock From coarse to fine: Robust hierarchical localization at large scale.
\newblock In {\em CVPR}, pages 12716--12725, 2019.

\bibitem{sarlin2020superglue}
Paul-Edouard Sarlin, Daniel DeTone, Tomasz Malisiewicz, and Andrew Rabinovich.
\newblock Superglue: Learning feature matching with graph neural networks.
\newblock In {\em CVPR}, pages 4938--4947, 2020.

\bibitem{schroff2015facenet}
Florian Schroff, Dmitry Kalenichenko, and James Philbin.
\newblock Facenet: A unified embedding for face recognition and clustering.
\newblock In {\em CVPR}, pages 815--823, 2015.

\bibitem{schuster2019sdc}
Ren{\'e} Schuster, Oliver Wasenmuller, Christian Unger, and Didier Stricker.
\newblock Sdc-stacked dilated convolution: A unified descriptor network for
  dense matching tasks.
\newblock In {\em CVPR}, pages 2556--2565, 2019.

\bibitem{tcl}
Yanqing Shen, Ruotong Wang, Weiliang Zuo, and Nanning Zheng.
\newblock Tcl: Tightly coupled learning strategy for weakly supervised
  hierarchical place recognition.
\newblock {\em Robot. Autom. lett.}, 7(2):2684--2691, 2022.

\bibitem{nordland}
Niko S{\"u}nderhauf, Peer Neubert, and Peter Protzel.
\newblock Are we there yet? challenging seqslam on a 3000 km journey across all
  four seasons.
\newblock In {\em IEEE Int. Conf. Robot. Autom. Worksh.}, page 2013, 2013.

\bibitem{teichmann2019detect}
Marvin Teichmann, Andre Araujo, Menglong Zhu, and Jack Sim.
\newblock Detect-to-retrieve: Efficient regional aggregation for image search.
\newblock In {\em CVPR}, pages 5109--5118, 2019.

\bibitem{pitts30k}
Akihiko Torii, Josef Sivic, Tomas Pajdla, and Masatoshi Okutomi.
\newblock Visual place recognition with repetitive structures.
\newblock In {\em CVPR}, pages 883--890, 2013.

\bibitem{selective}
Fusheng Wang, Jianhao Yan, Fandong Meng, and Jie Zhou.
\newblock Selective knowledge distillation for neural machine translation.
\newblock In {\em Int. Joint Conf. Natural Lang. Process.}, pages 6456--6466,
  2021.

\bibitem{transvpr}
Ruotong Wang, Yanqing Shen, Weiliang Zuo, Sanping Zhou, and Nanning Zheng.
\newblock Transvpr: Transformer-based place recognition with multi-level
  attention aggregation.
\newblock In {\em CVPR}, pages 13648--13657, 2022.

\bibitem{msls}
Frederik Warburg, Soren Hauberg, Manuel Lopez-Antequera, Pau Gargallo, Yubin
  Kuang, and Javier Civera.
\newblock Mapillary street-level sequences: A dataset for lifelong place
  recognition.
\newblock In {\em CVPR}, pages 2626--2635, 2020.

\bibitem{mobilesal}
Yu-Huan Wu, Yun Liu, Jun Xu, Jia-Wang Bian, Yu-Chao Gu, and Ming-Ming Cheng.
\newblock Mobilesal: Extremely efficient rgb-d salient object detection.
\newblock {\em IEEE TPAMI}, 2021.

\bibitem{zhang2021visual}
Xiwu Zhang, Lei Wang, and Yan Su.
\newblock Visual place recognition: A survey from deep learning perspective.
\newblock {\em Pattern Recognition}, 113:107760, 2021.

\bibitem{zhao2020knowledge}
Long Zhao, Xi Peng, Yuxiao Chen, Mubbasir Kapadia, and Dimitris~N Metaxas.
\newblock Knowledge as priors: Cross-modal knowledge generalization for
  datasets without superior knowledge.
\newblock In {\em CVPR}, pages 6528--6537, 2020.

\bibitem{zhou2}
Sanping Zhou, Fei Wang, Zeyi Huang, and Jinjun Wang.
\newblock Discriminative feature learning with consistent attention
  regularization for person re-identification.
\newblock In {\em ICCV}, pages 8040--8049, 2019.

\bibitem{zhou3}
Sanping Zhou, Jinjun Wang, Deyu Meng, Yudong Liang, Yihong Gong, and Nanning
  Zheng.
\newblock Discriminative feature learning with foreground attention for person
  re-identification.
\newblock {\em IEEE TIP}, 28(9):4671--4684, 2019.

\bibitem{zhou1}
Sanping Zhou, Jinjun Wang, Jiayun Wang, Yihong Gong, and Nanning Zheng.
\newblock Point to set similarity based deep feature learning for person
  re-identification.
\newblock In {\em CVPR}, pages 3741--3750, 2017.

\end{thebibliography}


\begin{thebibliography}{10}\itemsep=-1pt

\bibitem{netvlad}
Relja Arandjelovic, Petr Gronat, Akihiko Torii, Tomas Pajdla, and Josef Sivic.
\newblock {NetVLAD}: Cnn architecture for weakly supervised place recognition.
\newblock In {\em CVPR}, pages 5297--5307, 2016.

\bibitem{delg}
Bingyi Cao, Andre Araujo, and Jack Sim.
\newblock Unifying deep local and global features for image search.
\newblock In {\em ECCV}, pages 726--743, 2020.

\bibitem{cheng2020higherhrnet}
Bowen Cheng, Bin Xiao, Jingdong Wang, Honghui Shi, Thomas~S Huang, and Lei
  Zhang.
\newblock Higherhrnet: Scale-aware representation learning for bottom-up human
  pose estimation.
\newblock In {\em CVPR}, pages 5386--5395, 2020.

\bibitem{detone2018superpoint}
Daniel DeTone, Tomasz Malisiewicz, and Andrew Rabinovich.
\newblock Superpoint: Self-supervised interest point detection and description.
\newblock In {\em CVPR}, pages 224--236, 2018.

\bibitem{ge2020self}
Yixiao Ge, Haibo Wang, Feng Zhu, Rui Zhao, and Hongsheng Li.
\newblock Self-supervising fine-grained region similarities for large-scale
  image localization.
\newblock In {\em ECCV}, pages 369--386, 2020.

\bibitem{hausler2021patch}
Stephen Hausler, Sourav Garg, Ming Xu, Michael Milford, and Tobias Fischer.
\newblock {Patch-NetVLAD}: Multi-scale fusion of locally-global descriptors for
  place recognition.
\newblock In {\em CVPR}, pages 14141--14152, 2021.

\bibitem{dasgil}
Hanjiang Hu, Zhijian Qiao, Ming Cheng, Zhe Liu, and Hesheng Wang.
\newblock Dasgil: Domain adaptation for semantic and geometric-aware
  image-based localization.
\newblock {\em IEEE TIP}, 30:1342--1353, 2020.

\bibitem{li2018megadepth}
Zhengqi Li and Noah Snavely.
\newblock Megadepth: Learning single-view depth prediction from internet
  photos.
\newblock In {\em CVPR}, pages 2041--2050, 2018.

\bibitem{nord1}
Daniel Olid, Jos{\'e}~M F{\'a}cil, and Javier Civera.
\newblock Single-view place recognition under seasonal changes.
\newblock {\em arXiv preprint arXiv:1808.06516}, 2018.

\bibitem{paolicelli2022learning}
Valerio Paolicelli, Antonio Tavera, Carlo Masone, Gabriele Berton, and Barbara
  Caputo.
\newblock Learning semantics for visual place recognition through multi-scale
  attention.
\newblock In {\em Int. Conf. Image Anal. Process.}, pages 454--466, 2022.

\bibitem{unet}
Olaf Ronneberger, Philipp Fischer, and Thomas Brox.
\newblock U-net: Convolutional networks for biomedical image segmentation.
\newblock In {\em International Conference on Medical image computing and
  computer-assisted intervention}, pages 234--241, 2015.

\bibitem{sarlin2020superglue}
Paul-Edouard Sarlin, Daniel DeTone, Tomasz Malisiewicz, and Andrew Rabinovich.
\newblock Superglue: Learning feature matching with graph neural networks.
\newblock In {\em CVPR}, pages 4938--4947, 2020.

\bibitem{nordland}
Niko S{\"u}nderhauf, Peer Neubert, and Peter Protzel.
\newblock Are we there yet? challenging seqslam on a 3000 km journey across all
  four seasons.
\newblock In {\em IEEE Int. Conf. Robot. Autom. Worksh.}, page 2013, 2013.

\bibitem{pitts30k}
Akihiko Torii, Josef Sivic, Tomas Pajdla, and Masatoshi Okutomi.
\newblock Visual place recognition with repetitive structures.
\newblock In {\em CVPR}, pages 883--890, 2013.

\bibitem{msls}
Frederik Warburg, Soren Hauberg, Manuel Lopez-Antequera, Pau Gargallo, Yubin
  Kuang, and Javier Civera.
\newblock Mapillary street-level sequences: A dataset for lifelong place
  recognition.
\newblock In {\em CVPR}, pages 2626--2635, 2020.

\bibitem{xiao2018unified}
Tete Xiao, Yingcheng Liu, Bolei Zhou, Yuning Jiang, and Jian Sun.
\newblock Unified perceptual parsing for scene understanding.
\newblock In {\em ECCV}, pages 418--434, 2018.

\bibitem{zhao2017pyramid}
Hengshuang Zhao, Jianping Shi, Xiaojuan Qi, Xiaogang Wang, and Jiaya Jia.
\newblock Pyramid scene parsing network.
\newblock In {\em CVPR}, 2017.

\bibitem{adamw}
Hui Zhong, Zaiyi Chen, Chuan Qin, Zai Huang, Vincent~W Zheng, Tong Xu, and
  Enhong Chen.
\newblock Adam revisited: a weighted past gradients perspective.
\newblock {\em Frontiers of Computer Science}, 14(5):1--16, 2020.

\bibitem{ade20k}
Bolei Zhou, Hang Zhao, Xavier Puig, Sanja Fidler, Adela Barriuso, and Antonio
  Torralba.
\newblock Scene parsing through ade20k dataset.
\newblock In {\em Proceedings of the IEEE Conference on Computer Vision and
  Pattern Recognition}, 2017.

\end{thebibliography}
}
\end{CJK*}
\end{document}

\title{Distill Structural Knowledge by Weighting Samples \\ for Visual Place Recognition \\ Supplementary Material}
\author{Yanqing Shen, Sanping Zhou, Jingwen Fu, Ruotong Wang, Shitao Chen, and Nanning Zheng$^{*}$
\thanks{*This work was supported by the National Natural Science Foundation of China (Grant No. TODO).}
\thanks{Y. Shen, R. Wang, and N. Zheng are with the Institute of Artificial Intelligence and Robotics, Xi'an Jiaotong University, Xi'an, Shaanxi 710049, P.R. China;
{\tt\footnotesize qing1159364090, wrt072@stu.xjtu.edu.cn; spzhou@xjtu.edu.cn; nnzheng@mail.xjtu.edu.cn}}
\thanks{$^{*}$Corresponding author: Nanning Zheng}
}

\begin{CJK*}{UTF8}{gbsn}
\CJKindent

\maketitle

This supplementary material is structured as follows. 
In \sref{sec:1}, we explain the reasons for choosing the datasets in the main paper and present the detailed dataset configurations.
In \sref{sec:2}, we show some additional quantitative results and analysis on datasets, as well as results on the computation time. This section also contains several qualitative results. 
\sref{sec:3} provides the experimental details of methods, including detailed training settings of our method, ablation studies and the implementation of SOTA methods. 
Finally, \sref{sec:4} contains some additional ablation studies. 

\section{Detailed Dataset Configuration}
% todo
% 结构信息 contribution；pitts提升、不多；MSLS提升很多
% 有或没有 需求
% 数据集的选择
% 介绍的理由 和 主线
\label{sec:1}
% 简介；subset partitions for training, validation and testing；多少张图像；怎么测试；怎么挖掘样本；有哪些挑战

To facilitate an informed assessment of the results, we further detail datasets and the usage, which were briefly mentioned in Section 4.2 of the main paper. We evaluate our method on 4 key benchmark datasets.

\textbf{Mapillary Street Level Sequences (MSLS).}
MSLS~\cite{msls} is introduced to promote lifelong place-recognition research, and contains over 1.6 million images recorded in urban and suburban areas over 7 years.
Compared to other datasets, it covers the most comprehensive variation (dynamic objects, season, light, viewpoint, and weather), and we only evaluate the image-to-image task.
% the dataset is suitable for sequence-based methods, although
GPS coordinates and compass angles are provided for each image, and the ground truth corresponding to a query is the reference images located within 25m and 40$^{\circ}$ from the query. 
The dataset is divided into a training set, a public validation set and a withheld test set (MSLS challenge)\footnote{\href{https://codalab.lisn.upsaclay.fr/competitions/865}{https://codalab.lisn.upsaclay.fr/competitions/865}}.
In training, we define a distance $d_{qp}$ to represent the FOV overlap between query $q$ and positive $p$:
\begin{equation}
\label{eq:dis}
d_{q p}=\left\|\mathbf{x}_q-\mathbf{x}_p\right\|_2 / 25 + \left(\theta_q-\theta_p\right) / 40 < 1
\end{equation}
where $\mathbf{x}$ is the GPS coordinate and $\theta$ is the angle. \eref{eq:dis} ensures an overlapping area between a query and a positive.

\textbf{Nordland.} The Nordland dataset\cite{nordland} contains four timestamp-aligned image sequences recorded in four seasons, and hence it contains challenging appearance changes and different weather conditions while few viewpoint variations. %during a 728 km long train journey
We use the partitioned dataset \cite{nord1} containing 3450 images per sequence, with summer as reference and winter as query.
Same as \cite{hausler2021patch,nord1}, we also remove all black tunnels and times when the train is stopped.
Ground truth tolerance is set to 2 frames, that means that one query image corresponds to 5 reference images.

\textbf{Pittsburgh.} The Pittsburgh dataset \cite{pitts30k} contains 250k images derived from Google Street View panoramas.
The data is generated by 24 perspective images (two pitch and twelve yaw directions) at each place, which results in significant viewpoint variations, along with dynamic objects.
As only GPS information is available, the ground truths for evaluation are defined as reference images within 25m from the query.
In our experiments, we use the subset, Pitts30k, and the weakly supervised sample mining strategy \cite{netvlad} in training. It contains 30k database images and 24k queries, which are geographically divided into train/val./test sets.

% \textbf{Tokyo 24/7.} The Tokyo 24/7 dataset \cite{tokyo247} contains 76k database and 315 query images, where the queries taken using mobile phone cameras at three different times of the day ( daytime, sunset and night) while the database images were only taken at daytime from Google Street View as described above.
% The tolerance of ground truths for evaluation is 25m, same as Pittsburgh. Contrary to the Nordland and Pittsburgh datasets, Tokyo 24/7 includes nighttime images.

\section{Additional Results}
\label{sec:2}
% sota 对比
% \subsection{Quantitative results}

\begin{figure}
    \centering
    \includegraphics[width=0.9\linewidth]{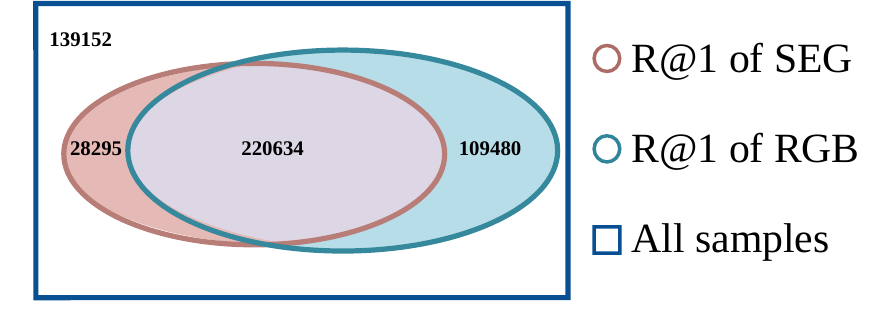}
    \caption{Visualization of sample distribution of MSLS training set. The numbers indicate the number of samples.}
    \label{sfig:ratio}
\end{figure}

\begin{figure*}
    \centering
    \includegraphics[width=0.9\linewidth]{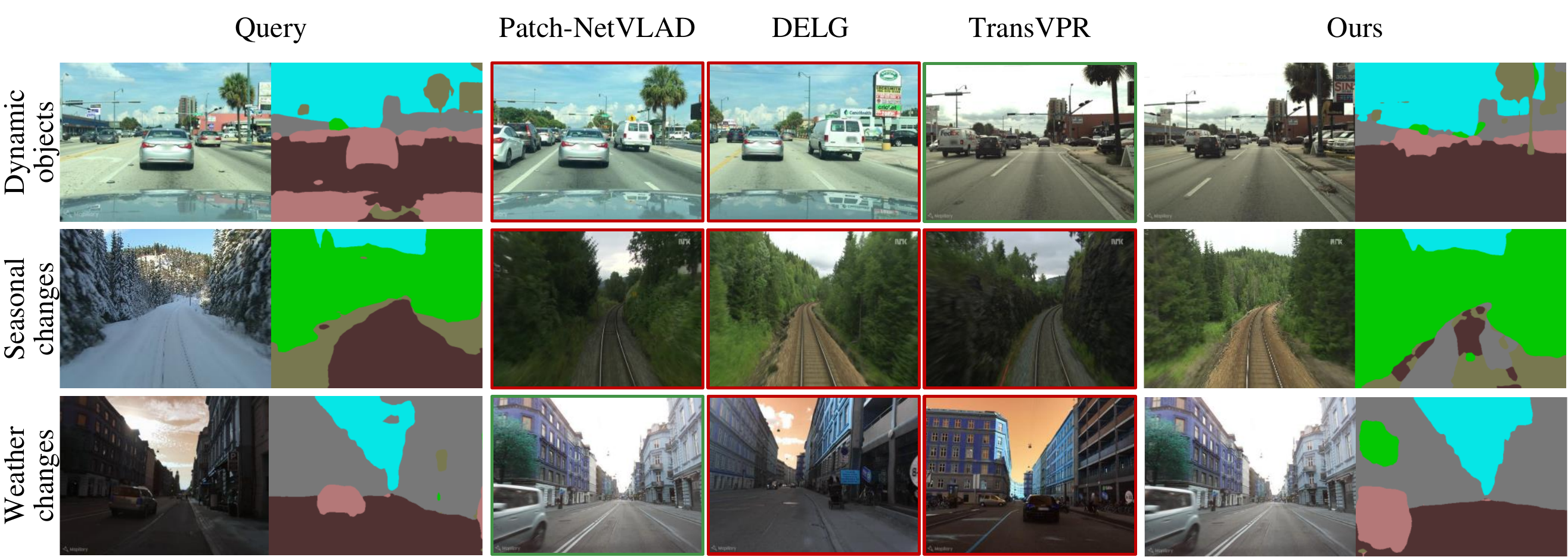}
    \caption{\textbf{Qualitative Results.} In these examples, our method successfully retrieves the matching reference images. For other methods, red borders indicate false matches and green borders indicate correct matches.}
    \label{fig:vis}
\end{figure*}
\textbf{Complementarity of RGB and SEG.} In the main paper, we mentioned that there is a sample-level complementarity between RGB images and segmentation images for VPR. 
We calculate the Recall$@1$ of seg-branch in the query sets with correct/wrong predictions at top-1 ranking list of rgb-branch, respectively, and we swap branches and do the calculation again.
The specific results on the MSLS training set are as shown in \fref{sfig:ratio}.

\begin{figure}
    \centering
    \includegraphics[width=0.8\linewidth]{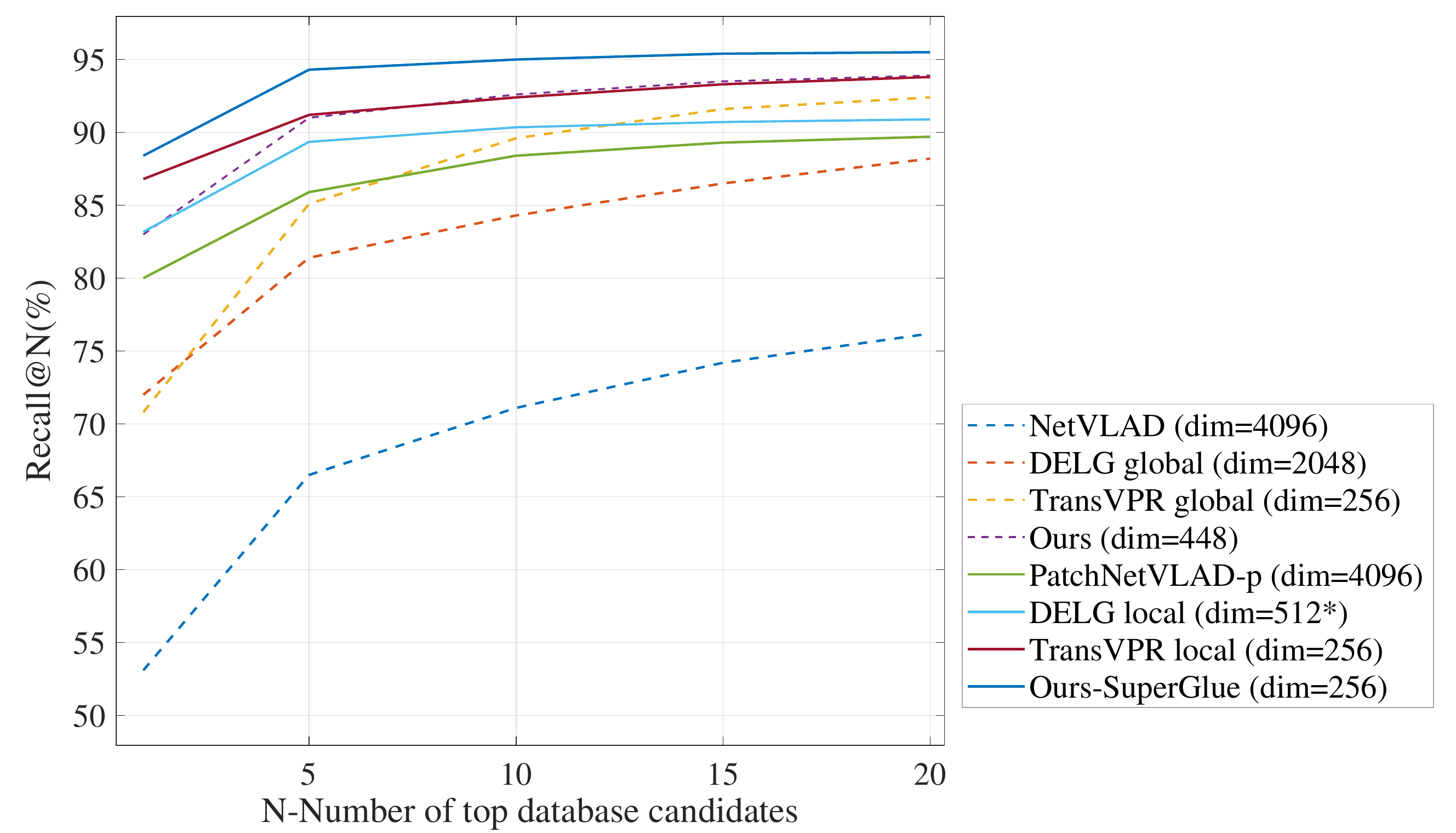}
    \vspace{-0.2cm}
    \caption{\textbf{Comparison with state-of-the-art on MSLS val. set.} We show the comparison of Recall$@N$ performance with other methods. Results w/o re-ranking are depicted in dotted line, while results with re-ranking are depicted in solid line. * indicates unofficially reproduced results, and details are in \sref{sub:baseline}.}
    \label{sfig:msls_compare}
\end{figure}

\begin{table}
  \centering
\caption{Performance of selective distillation with different samples on Nordland dataset. None refers to the \textit{rgb-branch} without distillation and All refers to non-selective distillation.}
\label{stab:group_abl}
  \scalebox{0.65}{
 \renewcommand{\arraystretch}{1.2}
\begin{tabular}{c|c|ccccc}
\toprule
\multirow{2}{*}{\begin{tabular}[c]{@{}c@{}}Group for \\ distillation\end{tabular}} & \multirow{2}{*}{\begin{tabular}[c]{@{}c@{}} Sample\\ratio \end{tabular}}&\multicolumn{5}{c}{Nordland}  \\
\cline{3-7}
& & R@1      & R@5     & R@10    & R@15       & R@20    \\
\midrule
None & 0\% & 48.3 & 69.2 & 76.1 & 80.6 & 83.0\\
\textbf{All} & 100\% & \underline{55.4} & 71.9 & 76.9 & 79.4 & 80.7\\ % 0707
\midrule
${\mathcal{D}_{1}}$ & 4.11\% & 51.2 & 69.1 & 75.9 & 79.8 & 82.5\\ % 0803
${\mathcal{D}_{2}}$ & 50.67\%  & 52.5 & 72.1 & 79.8 & \underline{83.5} & \underline{85.8}\\ % 0806
${\mathcal{D}_{3}}$ & 14.77\%  &42.4& 62.5 & 70.8 & 75.1 & 78.4 \\ % 0524
${\mathcal{S}_{1}}$& 69.55\% & {54.8} & \underline{74.0}& \underline{81.1} & 83.2 & 84.8\\ % 0813
${\mathcal{S}_{2}}({\mathcal{D}_{4}})$ & 30.45\% &33.0 & 54.4 & 64.2 & 69.5 & 72.8 \\ % 0518
${\mathcal{R}_{1}}$ & 73.24\% &55.2 & 73.7 & 79.6 & 82.6 & 84.0\\ % 0806
${\mathcal{R}_{2}}$ & 26.76\% &51.0 & 67.4 & 73.0 & 75.4 & 77.8\\ % 0518
\midrule
\textbf{Ours} & 69.55\% & \textbf{56.1} & \textbf{75.5} & \textbf{82.9} & 86.2 & \textbf{88.3}\\
\bottomrule
\end{tabular}}
\end{table}

\begin{figure}
    \centering
    % \captionsetup[subfloat]{labelsep=none,format=plain,labelformat=empty}
    \subfloat[Global methods]{
    %   \label{fig:seg_val}
      \includegraphics[width=0.4\linewidth]{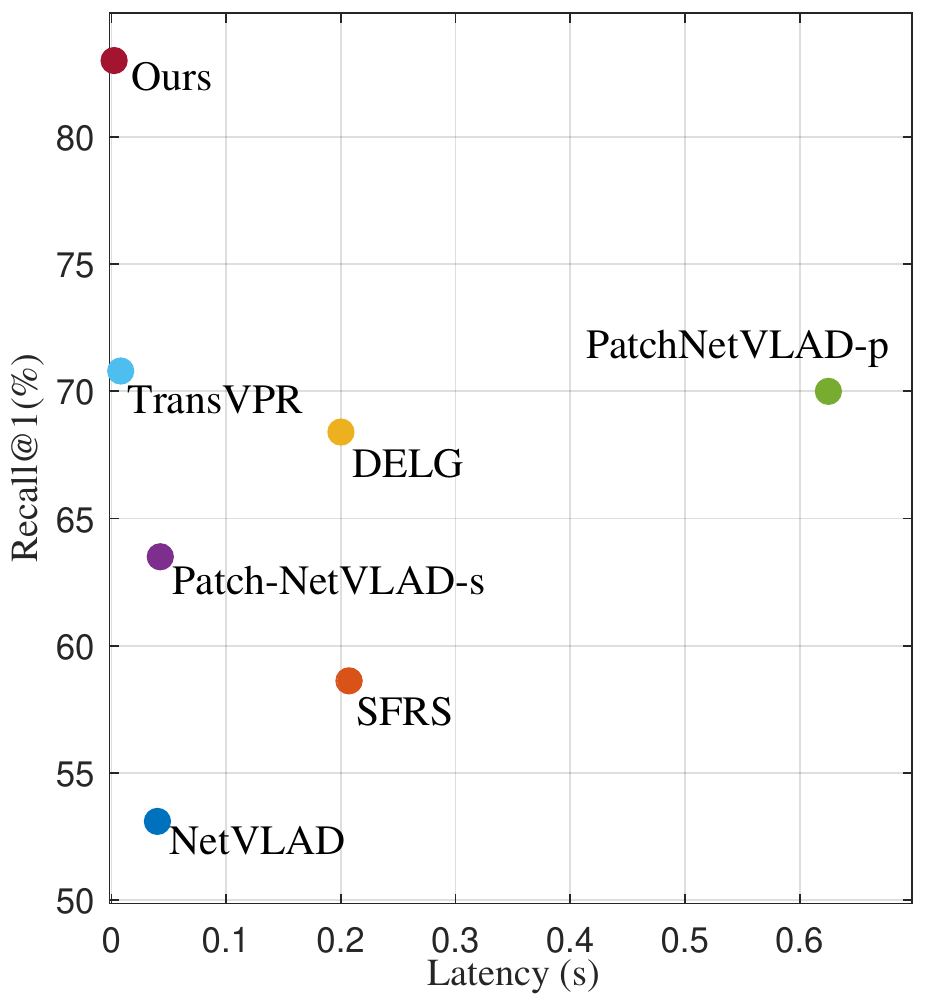}}
    \subfloat[Two-stage methods]{
    %   \label{fig:seg_test}
      \includegraphics[width=0.4\linewidth]{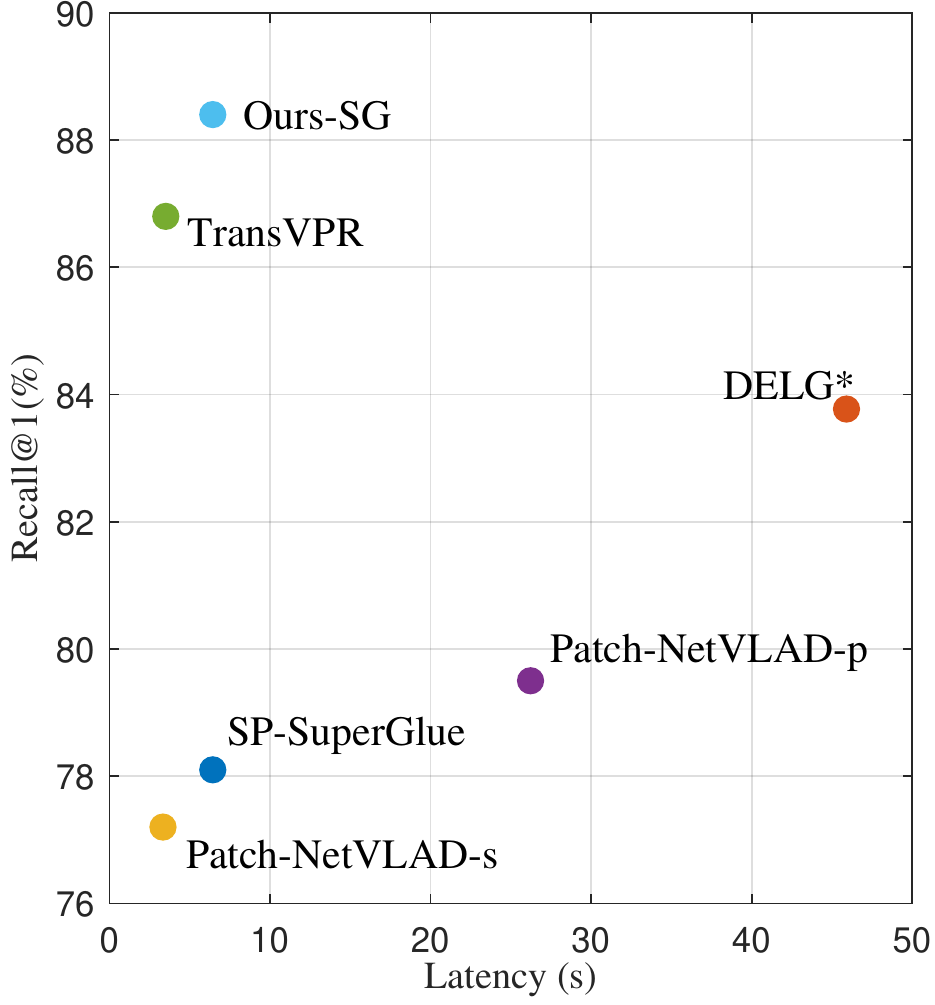}}
      \vspace{-0.2cm}
    \caption{The Recall@1 scores on MSLS validation set are shown on the y-axis, and the x-axis represents the accumulated time of feature extraction and feature matching.}
    \label{sfig:time}
\end{figure}

\begin{table}
  \centering
\caption{Performance of \textit{seg-branch} with different number of clustered classes. Note that $\dagger$ indicates another weighted encoding.}
\label{stab:number_abl}
  \scalebox{0.7}{
 \renewcommand{\arraystretch}{1.2}
\begin{tabular}{c|ccccc}
\toprule
\multirow{2}{*}{\begin{tabular}[c]{@{}c@{}}Clustered  \\ classes\end{tabular}} & \multicolumn{5}{c}{Nordland} \\
\cline{2-6}
&R@1  & R@5  & R@10   & R@15   & R@20 \\
\midrule
3 & 40.4 & 63.6 & \underline{73.6} & 78.0 & 81.1\\
6 & \underline{41.1} & \underline{63.8} & {73.2} & \underline{78.6} & \underline{81.6}\\
150& 25.1 & 41.4 & 50.4 & 55.3 & 58.9\\
\midrule
6 $^\dagger$ & 34.5 & 53.0 & 61.3 &65.9 &69.1\\
6 (Ours) & \textbf{56.1} & \textbf{75.5} & \textbf{82.9} & \textbf{86.2} & \textbf{88.3} \\
\bottomrule
\end{tabular}}
\end{table}

\textbf{Qualitative results.}
\fref{fig:vis} illustrates example matches with challenging conditions, such as viewpoint changes, occlusions caused by dynamic objects, and seasons. 
In these examples, other methods show a tendency to retrieve images with similar appearance as the query. Especially in the case of MSLS, the color tone and the vehicle ahead of retrievals of Patch-NetVLAD and DELG are consistent with query.
Ours can successfully retrieve images based on structural information, paying more attention to the spatial information of static objects in the background.

\textbf{Additional recall plots.}
Table 1 in the main paper shows the Recall$@N$ performance on the benchmark datasets. More intuitively, \fref{sfig:msls_compare} shows the detailed Recall$@N$ performance for the MSLS validation dataset.

\textbf{Efficiency-accuracy.} Table 2 in the main paper shows the feature extraction, feature matching time and storage required to process each query. In \fref{sfig:time}, we show the the accumulated time of feature extraction and matching as well as their performances on MSLS validation set. Ours achieves the best trade-off between accuracy and efficiency.

\textbf{Ablations on other datasets.}
In \tref{stab:group_abl} and \tref{stab:number_abl}, we provide more results on Nordland dataset for ablation experiments discussed in the main paper.

\section{More Implementation Details}
\label{sec:3}
% 对应的框架图？
\subsection{Training Details}

\textbf{Ours.} 
To obtain segmentation images offline, we use ADE20K \cite{ade20k}\footnote{The largest open-source dataset for semantic segmentation and scene parsing, similar to the distribution of VPR datasets.} and PSPNet\cite{zhao2017pyramid} and the open-source code-base \footnote{\href{https://github.com/CSAILVision/semantic-segmentation-pytorch}{https://github.com/CSAILVision/semantic-segmentation-pytorch}} with configuration file of ``ade20k-resnet50dilated-ppm\_deepsup''.
MobileNetV2 is initialed with the pre-trained weights on ImageNet \footnote{\href{https://download.pytorch.org/models/mobilenet_v2-b0353104.pth}{https://download.pytorch.org/models/mobilenet\_v2-b0353104.pth}}. 
MobileNet-L in seg-branch refers to the implementation of depth-stream in MobileSal\footnote{\href{https://github.com/yuhuan-wu/MobileSal}{https://github.com/yuhuan-wu/MobileSal}}.

In the first learning stage, rgb-branch and seg-branch are fine-tuned with the whole backbone using initial lr=0.001, where rgb-branch starts with a pre-trained model on ImageNet and seg-branch starts with random parameters.
In the second learning stage, backbone also starts with a pre-trained model on ImageNet and is fine-tuned as a whole with initial lr=0.0001.
We also attempted to continue the second stage on the basis of pre-trained network in the first training stage, but the difference is not significant.

Models are all optimized by AdamW optimizer \cite{adamw} with 0.0001 weight decay and cosine learning rate decay schedule, and $m$ in VPR loss is 0.1.
The network which yields the best recall@5 on the val. set is used for testing.

For running SuperGlue network with SuperPoint based on our global retrieval, the details are shown in \sref{sub:baseline}.

\textbf{Concat-input.}
We concatenate RGB image and encoded segmentation label map in channel $C$ as input, where the $C$ of the first layer changes from 3 to 9 compared with \textit{rgb-branch}.
The model is fine-tuned with the whole backbone and initial lr=5e-5. The dim of global features is 448.

\textbf{Concat-feat.}
Two separate networks, same as the two branches in the first stage, are used to extract features separately. Then the two features are concatenated, followed by a $L_2$ normalization step, as final global features to build loss function.
The two models are fine-tuned with the whole backbone and initial lr=0.00005. The final loss is the direct sum of the two losses.
The dim of global features is 928.

\textbf{Multi-task.}
The implementation of decoder refers to U-Net~\cite{unet} \footnote{\href{https://github.com/milesial/Pytorch-UNet}{https://github.com/milesial/Pytorch-UNet}}.
The model is fine-tuned with the whole backbone and initial lr=0.0001.
The final loss is the sum of the vanilla VPR loss and the weighted cross-entropy loss, where the weight of cross-entropy is 0.1.
The dim of global features is 448.

It is worth noting that the results of the Multi-task are not as good as expected, and the training process has high requirements for parameter tuning. 
Furthermore, compared with implicit supervision in the form of encoder-decoder, knowledge distillation is more direct and interpretable for enhancing structural information in features.

\subsection{Implementation Details of Baselines}
\label{sub:baseline}
\textbf{NetVLAD\cite{netvlad}.}  We use the pytorch implementation\footnote{\href{https://github.com/Nanne/pytorch-NetVlad}{https://github.com/Nanne/pytorch-NetVlad}} and its released model trained on Pitts30k
training set with VGG-16 backbone. Note that this method does not resize the image.

\textbf{SFRS\cite{ge2020self}.} 
This work proposes a self-supervised method with image-to-region similarities to fully explore the potential of difficult positive images alongside their sub-regions. 
We use the official implementation\footnote{\href{https://github.com/yxgeee/OpenIBL}{https://github.com/yxgeee/OpenIBL}} and the released model trained on Pitts30k training set. 

\textbf{SP-SuperGlue\cite{detone2018superpoint,sarlin2020superglue}.} SuperGlue trains a neural network that matches two sets of local features by jointly finding correspondences and rejecting non-matchable points. The implementation of SP-SuperGlue in our main paper includes: using NetVLAD for global retrieval, then extracting SuperPoint local features, and applying SuperGlue to identify matches and to re-rank candidates. We use the the official implementation \footnote{\href{https://github.com/magicleap/SuperGluepre-trainedNetwork}{https://github.com/magicleap/SuperGluepre-trainedNetwork}} and choose the pre-trained outdoor weights on MegaDepth dataset\cite{li2018megadepth}.

\textbf{Patch-NetVLAD\cite{hausler2021patch}.} This work derives patch-level features from NetVLAD residuals. We use the official implementation \footnote{\href{https://github.com/QVPR/Patch-NetVLAD}{https://github.com/QVPR/Patch-NetVLAD}} for speed-focused and performance-focused configurations in our main paper. 
Following the original paper, the model trained on Pitts30k is used for urban imagery (Pittsburgh)
% and Tokyo datasets)
, and the model trained on MSLS is used for all other conditions.

\textbf{DELG\cite{delg}.} This work unifies global and local features into a single deep model. We refer to the pytorch implementation of two models \footnote{\href{https://github.com/feymanpriv/DELG}{https://github.com/feymanpriv/DELG}}\footnote{The results in Table 2 in the main paper are the best of the two models.} and change the extraction of global features (dim=2048) to the way in the original paper: For global features, we use 3 scales; $L_2$ normalization is applied for each scale independently, then the three global features are average-pooled, followed by another $L_2$ normalization step.
For local features, all the reproduced results are from the provided model, in which dim is 512 and is different from the original paper (dim=128).

\textbf{TransVPR.} This work proposes a holistic model based on vision Transformers, which can aggregate task-relevant features. We use the official implementation\footnote{\href{https://github.com/RuotongWANG/TransVPR-model-implementation}{https://github.com/RuotongWANG/TransVPR-model-implementation}}. Same to Patch-NetVLAD, it finetuned the model on MSLS training set and Pitts30k training set.

\section{Additional Ablation Studies and Analysis}
\label{sec:4}

\textbf{Prior weights for encoding.}
% 能看出比全1是好的，大于、小于1的倾向是对的；但具体的最优值并没有遍历；To find the optimal weights of each semantic label, we can use grid search or learning to find the configuration that performs best on the training set.
% sample pair 不包含negative;  but it is unrealistic in training due to the huge number of possible tuples.So the optimal way is to traverse and partition all training tuples,
%todo 属于对未来工作的分析？
% todo 聚类数量
% % 6层中虽然只有3种取值 但也是不同的
% % 针对既定的6类
% 不区分地处理不同的label初值
% 其他的权重-比如相反的 肯定不对 initialization with 1 and 0.
% todo 需要展示更多的情况吗，比如更多的值？自行聚合类别？
Limited to the length of the article and the research content, in Section 4.5 of the main paper, we only manually set three different weighting cases and choose the best of the three, without finding the optimal one.
% 3种指的是全1，ours，opposite
Figure 6 indicates that the value of static buildings should be larger than that of dynamic objects, which is accorded with our intuition.
This weighted attempt provides the possibility for follow-up research, and further advancements can be done, such as using grid search or learning methods to obtain the weights through iterative optimization.

% According to the situation of the dataset and combined with human experience
% 植被其实很关键 季节
% pitts里用的应该是另一种权重方案啊

\textbf{Weighting function.}
For the weight function Eq. (4) given in the main paper, we make the following supplementary explanations.
In main paper, we attempted to apply different constant weights to different groups according to the performance of separate group in Table 4. 

\begin{figure}
    \centering
    \includegraphics[width=0.8\linewidth]{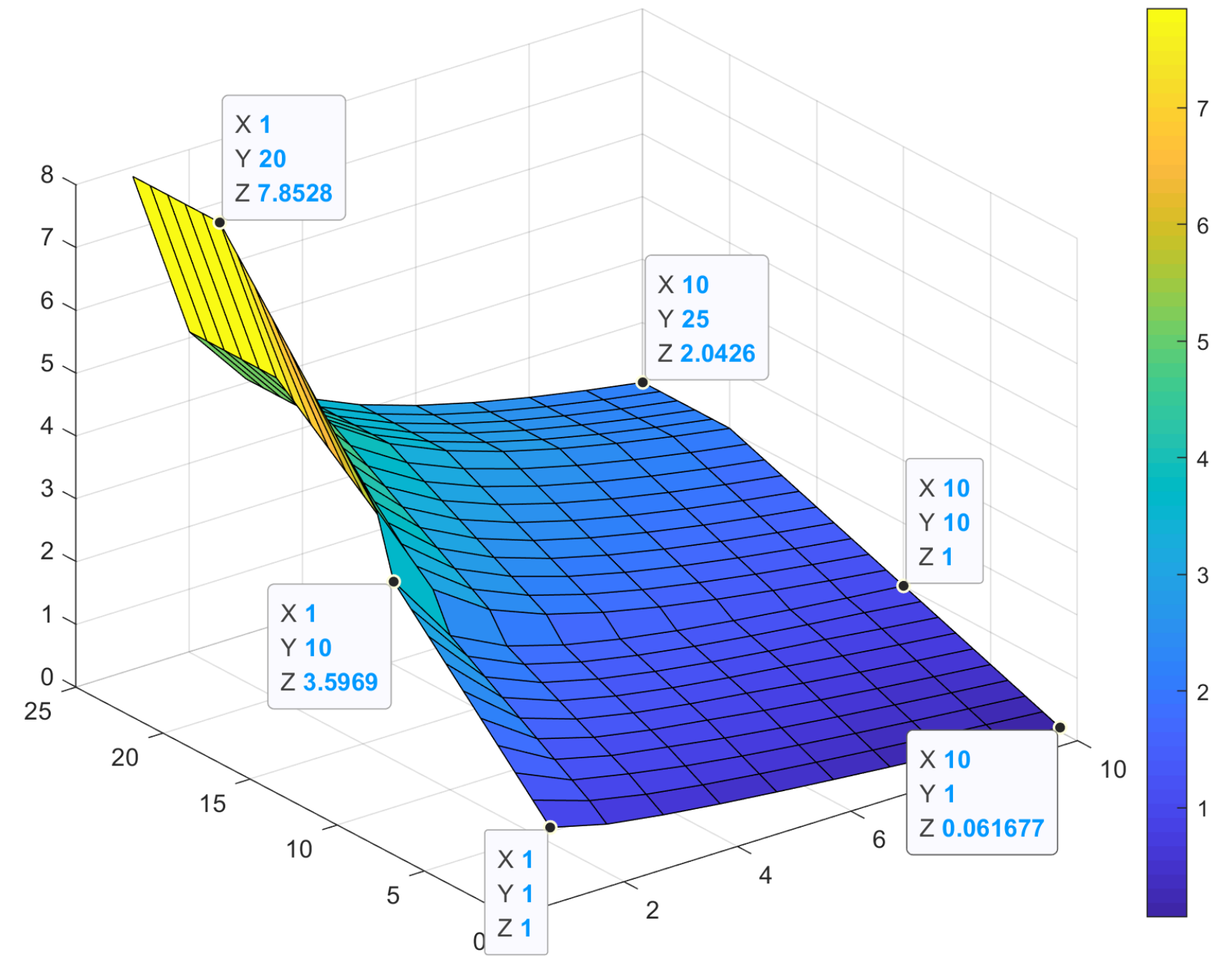}
    \caption{Visualization of the weight function. Some weights of  demarcation points are marked. $X$ and $Y$ mean the recall rankings of samples of seg-branch and rgb-branch, same as the main paper.}
    \label{sfig:weight}
\end{figure}

\begin{table}
  \centering
\caption{Performance of weighted distillation with different weights. The weights correspond to $\mathcal{D}_{1}$-$\mathcal{D}_{2}$-$\mathcal{D}_{3}$-$\mathcal{D}_{4}$.}
\label{stab:weight_abl}
  \scalebox{0.7}{
 \renewcommand{\arraystretch}{1.2}
\begin{tabular}{c|ccc|ccc}
\toprule
\multirow{2}{*}{\begin{tabular}[c]{@{}c@{}}Weight\end{tabular}} & \multicolumn{3}{c|}{MSLS val} & \multicolumn{3}{c}{MSLS challenge} \\
\cline{2-7}
& R@1      & R@5     & R@10    & R@1       & R@5       & R@10      \\
\midrule
GP-D(1-4-8-0) &  {80.3} & 86.9 & 90.2 & {61.2} & 76.1 & 81.2\\ %
GP-D(1-2-4-0) &  {80.6} & 87.6 & 90.4 & {61.5} & 76.3 & 81.1\\ % 
GP-D(4-2-1-0) &  \underline{81.2} & \textbf{89.3} & \underline{91.4} & \underline{61.7} & \underline{78.3} & \textbf{82.5}\\ %0808
GP-D(8-4-1-0) &  \textbf{82.2} & \underline{88.9} & \textbf{91.5} & \textbf{62.3} &\textbf{78.9} & \underline{82.4}\\ % 0719
\bottomrule
\end{tabular}}
\end{table}
% 反向分段量化取值【提升越高，权值越大】
% 差别不是很大 所以选择了8 4 1这一组
% 4 5这些数字
% 实际上还做了几组离散数值的实验：为了给连续权重函数的设计提供数值参考
In fact, we also tried other constant weights, as shown in \tref{stab:weight_abl}.
The results show that GP-D(8-4-1-0) and GP-D(4-2-1-0) has a more reasonable weight distribution than others, which proves our prediction of the importance of different groups: the greater the performance improvement when participating in distillation alone, the higher the weight.
Moreover, the difference between GP-D(8-4-1-0) and GP-D(4-2-1-0) is small, and this experiment is mainly for providing a numerical reference for the design of our weight function.
Therefore, we finally choose GP-D(8-4-1-0) as the reference.

Based on reference constant weights, the specific function design includes the following consideration:
\begin{itemize}
    \item $\varphi$ cannot be negative;
    \item The non-zero part of $\varphi$ should be proportional to $y-x$ and inversely proportional to $x$;
    \item The value of $\varphi$ cannot be too large;
    \item The partial derivatives should be different for 3 groups.
\end{itemize}

The prototype of the function can be denoted as $\frac{f(y-x)}{g(x)}$, where $f(\cdot)$ and $g(\cdot)$ are monotonically increasing functions. Considering $x$ should play an more important role in weights than $y-x$, we choose liner for $f(\cdot)$ and natural logarithm for $g(\cdot)$ (see \fref{sfig:weight}).

Throughout the design process, we did not perform rigorous tuning of the parameters, but simply chose representative design to demonstrate our insight and motivation.
In \tref{tbl:function}, we show more ablation experiments by replacing $\frac{y-x}{ln(x+1)}$ with $\frac{y-x}{x}$ in (4).
Combined with the performance of distillation with fixed weights in Table 3, it can be seen that the function performs better than the discrete fixed weights and prototype function, showing the advantages of (4).

\begin{table}[ht]
\footnotesize
  \caption{Comparisons of functions.}
  \label{tbl:function}
  \centering
  \scalebox{0.9}{\begin{tabular}{c|c|c}
  \toprule
  $\frac{y-x}{x}$ & MSLS val (81.2/89.5/92.2) & MSLS challenge (64.2/79.1/83.2)\\
  \midrule
  $\frac{y-x}{ \ln (x+1)}$ & MSLS val (83.0/91.0/92.6) & MSLS challenge (64.5/80.4/83.9)\\
  \bottomrule                  
  \end{tabular}}
\end{table}

\textbf{Sensitivity to hyper-parameters.}
In Section 4.4 of the main paper, we have performed ablation experiments on the most important hyper-parameters.
We further evaluate the sensitivity of our model to changes in the other two hyper-parameters: $N_t$ and $N_m$ in our group partition strategy.
% 画图：Nt选择1、5、10、15、20、100【合适范围内差别不大】
% 选定10，Nm选择15、20、25

\begin{figure}
    \centering
    \includegraphics[width=0.7\linewidth]{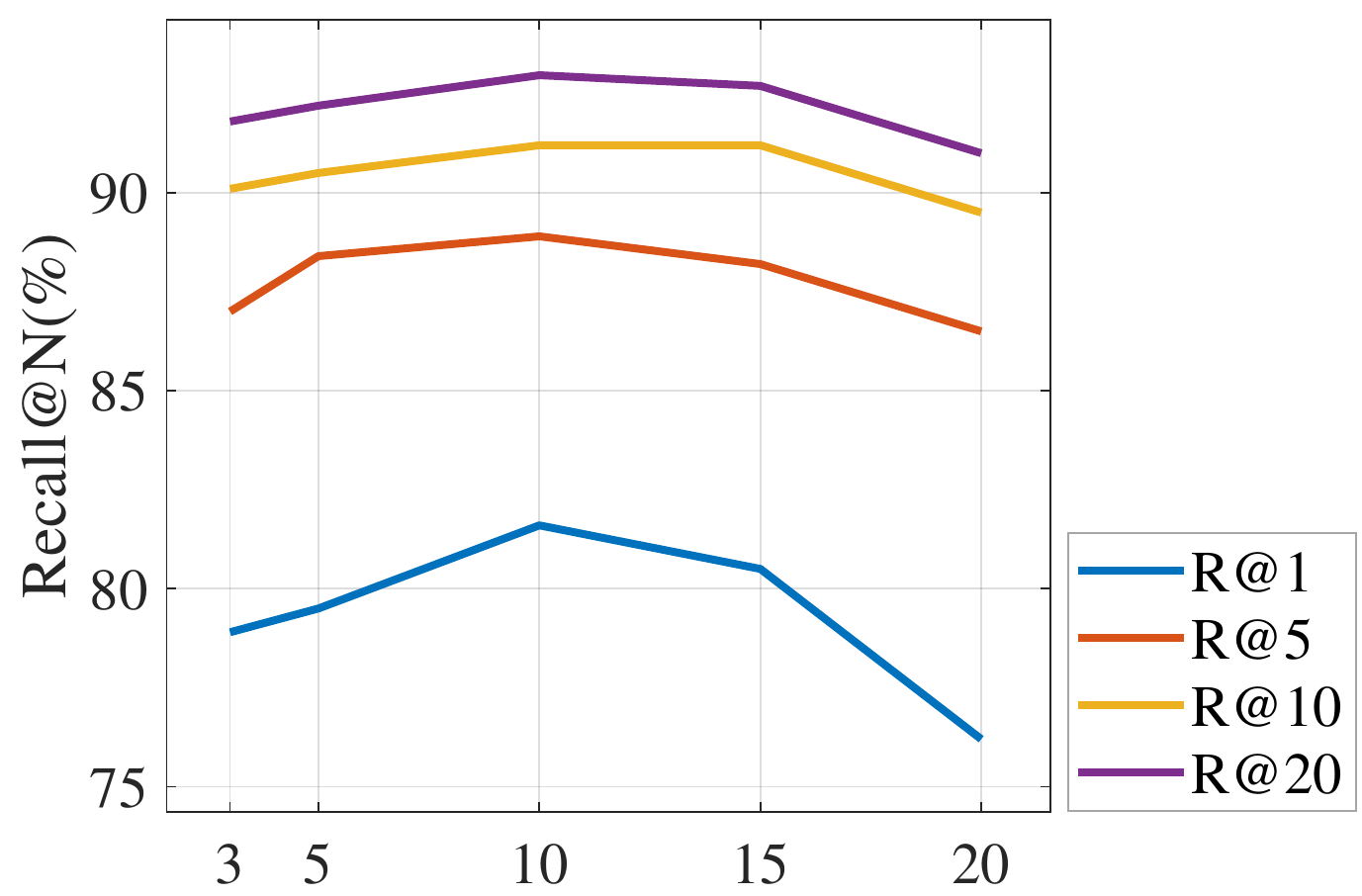}
    \caption{Ablation experiments on the recall performance of StructVPR with different $N_t$.}
    \label{sfig:group}
\end{figure}

Here we perform unweighted selective distillation with the experimental setup of GP-S, that is, on samples belong to $\mathcal{S}_1$.
The results are shown in \tref{stab:group} and \fref{sfig:group}.
It can be seen that within the appropriate range of 5-15, the performance is relatively close and we select 10 in the main paper.

       %  3 & 78.9 & 87.0&90.1 &91.8\\0726
       % 5  & 79.5 &88.4 & 90.5 & 92.2\\0726
       % 10 & 81.6 & 88.9 & 91.2 &92.97\\
       % 15 & 80.5 & 88.2 &91.2&92.7\\0719
       % 20 & 76.2 & 86.5 & 89.5 & 91.0\\0719

\begin{table}
\centering
    \caption{The sample ratio corresponding to different $N_t$.}
    \label{stab:group}
  \scalebox{0.8}{
    \begin{tabular}{c|ccccc}
    \toprule
    $N_t$ &3 & 5 &10 &15 & 20\\ 
    \midrule
       %  3 & 2.62 & 50.44 & 7.69 & 39.25\\
       % 5  & 3.7 & 50.13 & 10.93 & 35.24\\
       % 10 & 4.11 & 50.67 & 14.77 & 30.45\\
       % 15 & 5.7 & 49.5 & 16.91 & 27.89\\
       % 20 & 6.92 & 48.51 &18.39 & 26.18\\
      $\mathcal{S}_1$ &60.75\% &64.76\% &69.55\% & 72.11\%&73.82\%\\
      $\mathcal{S}_2$ &39.25\% &35.24\% &30.45\% & 27.89\%&26.18\%\\
    \bottomrule
    \end{tabular}}
\end{table}

% Nt其实是知识是否可以接受的分界线；Nm则是限制权重范围
% 那关于超参 是不是强调正文里的SLME

After $N_t$ is set as 10, $N_m$ is mainly used to limit the weight range. %todo

\textbf{Sensitivity to Segmentation Models.}
% 有些工作在引入segmentation或depth的时候，为了使用精准的外源信息，从虚拟数据集训练再域适应到真实数据集上；而我们则直接采用开源模型得到数据，并通过提出的方法完成性能的提升，
% todo 写在这个节里面，好像无法避免segmentation？
% \textcolor{blue}
{In order to use accurate semantic information, some previous works\cite{dasgil,paolicelli2022learning} use synthetic virtual datasets for training, and then generalize to real-world datasets through domain adaptation.}
In the main paper, we use an open-source semantic segmentation model to obtain segmentation images, which greatly reduces the implementation costs and training difficulty.

We also use another two commonly used semantic segmentation models (UPerNet\cite{xiao2018unified} and HRNetV2\cite{cheng2020higherhrnet}) for training to assess the sensitivity of StructVPR to the segmentation models.
We use the code-base\footnote{\href{https://github.com/CSAILVision/semantic-segmentation-pytorch}{https://github.com/CSAILVision/semantic-segmentation-pytorch}} with configuration file of ``ade20k-resnet50-upernet.yaml'' and ``ade20k-hrnetv2.yaml''.
% HRNetV2从精确度上更高;可能最终也更好诶
% 如果真的，是说我们有意选择了中间？还是更新main的结果？
% 对很多模型都具有兼容性

\begin{figure}
    \centering
    \includegraphics[width=\linewidth]{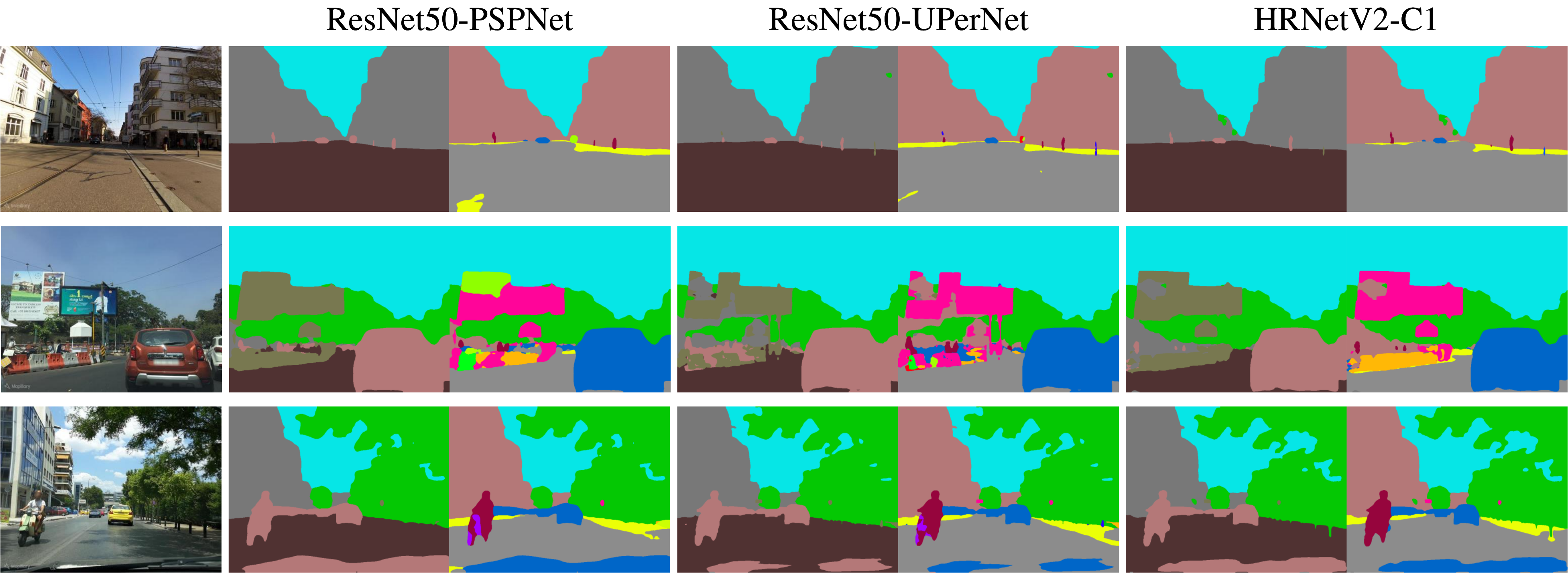}
    \caption{Examples of three semantic segmentation models with 6 classes and 150 classes.}
    \label{sfig:segmodel}
\end{figure}

As shown in \fref{sfig:segmodel}, HRNetV2 is slightly better than PSPNet and PSPNet is slightly better than UperNet with fewer holes in the case of 150-class segmentation, while in the case of 6-class segmentation, the difference among the three models becomes smaller.

\begin{table}
  \centering
\caption{Performance of \textit{seg-branch} and StructVPR with different segmentation models. PSP is for ResNet50-PSPNet (main paper), UPer is for ResNet50-UPerNet, and HR is for HRNetV2-C1.}
\label{stab:seg_abl}
  \scalebox{0.585}{
 \renewcommand{\arraystretch}{1.2}
\begin{tabular}{c|ccc|ccc|ccc}
\toprule
\multirow{2}{*}{Method} & \multicolumn{3}{c|}{MSLS val}& \multicolumn{3}{c|}{MSLS test} & \multicolumn{3}{c}{Nordland} \\
\cline{2-10}
&R@1  & R@5  & R@10 &R@1  & R@5  & R@10  &R@1  & R@5  & R@10\\
\midrule
SEG-UPer & 66.2&\textbf{80.7}&83.8&41.5&58.7&64.6&39.7&62.4&71.5\\
SEG-HR & 65.2&80.4&\textbf{84.1}&\textbf{44.0}&\textbf{59.5}&64.9&\textbf{44.6}&\textbf{65.1}&72.5\\
\textbf{SEG-PSP} &\textbf{67.7}&80.0&83.1&43.4&58.9&\textbf{65.8} & 44.4&64.8&\textbf{72.7}\\
\midrule
StructVPR-UPer & 82.4&90.3&\textbf{92.8}&63.4&78.8&82.7&57.2&75.6&82.4\\
StructVPR-HR &81.62 & 90.41&91.9 &60.8& 79.3&83.0 &\textbf{59.3}&\textbf{77.7}&\textbf{84.7} \\
\textbf{StructVPR-PSP} &\textbf{83.0} & \textbf{91.0} & {92.6} & \textbf{64.5} & \textbf{80.4}& \textbf{83.9}&56.1&75.5&82.9\\
\bottomrule
\end{tabular}}
\end{table}

\tref{stab:seg_abl} shows that StructVPR is compatible with many models and is little affected by segmentation models.
It is worth noting that due to time constraints, the results of ``StructVPR-HR'' and ``StructVPR-UPer'' are only the best performance among the current checkpoints, and we can provide the latest results later.
% todo
This also means that StructVPR can achieve excellent performance without relying on semantic annotations ground truth.
This is expected since the structural information we extract does not rely on completely accurate pixel-level segmentation, but more on spatial relative positional relationships.
Moreover, the clustering operation in SLME also makes StructVPR less sensitive to segmentation results.

\textbf{Analysis.} StructVPR achieves better performance and maintains a low inference cost without re-ranking. Compared to rgb-branch, StructVPR does have more costs in training due to the large amount of training set. Nevertheless, compared to model training, annotation and group partition are not expensive and mostly one-time efforts.
At last, considering the robustness of StructVPR to segmentation label map after SLME, we can seek smaller models or reduce the resolution to reduce costs of computing SEG images.

\bibliographystyle{ieee_fullname}
\bibliography{ref}
\end{CJK*}